\documentclass[11pt]{article}

\usepackage{acl}
\usepackage{times}
\usepackage{latexsym}
\usepackage[T1]{fontenc}
\usepackage[utf8]{inputenc}
\usepackage{microtype}
\usepackage{inconsolata}
\usepackage{graphicx}
\usepackage{booktabs}
\usepackage{multirow}
\usepackage{amsmath}
\usepackage{float}
\usepackage{amssymb}
\usepackage{tcolorbox}
\usepackage{listings}
\usepackage{cleveref}
\usepackage{xcolor}

\title{Self-Calibrating Language Models via Test-Time Discriminative Distillation}

\author{
 \textbf{Mohamed Rissal Hedna},
 \textbf{Jan Strich},
 \textbf{Martin Semmann,
 \textbf{Chris Biemann}
 }\\
 Hub of Computing and Data Science (HCDS) \\
 University of Hamburg, Germany \\
 \\
 \small{
   \textbf{Correspondence: \texttt{\{first\_name\}.\{last\_name\}@uni-hamburg.de}}
 }
}

\begin{document}
\maketitle

\begin{abstract}
Large language models (LLMs) are systematically overconfident: they routinely express high certainty on questions they often answer incorrectly.
Existing calibration methods either require labeled validation data, degrade under distribution shifts, or incur substantial inference costs.
Recent work has shown that LLMs already contain a better-calibrated signal than the one they verbalize: the token probability of "True" when the model is asked "Is this answer correct?" ($P(\text{True})$) consistently outperforms their stated confidence, a gap that is theoretically grounded as generative error is lower-bounded by roughly twice the corresponding discriminative error.
We introduce \textbf{SECL} (\textbf{SE}lf-\textbf{C}alibrating \textbf{L}anguage Models), a test-time training (TTT) pipeline that exploits this gap as label-free self-supervision, requiring no labeled data or human supervision. SECL adapts only when the input distribution shifts, training on just 6--26\% of the question stream at lower cost than the baseline it distills from.
Across four small language models from three model families and four diverse domains, SECL reduces Expected Calibration Error (ECE) by 56--78\%, outperforming its own supervision signal and matching or outperforming recent inference-time methods.
SECL is the first method to apply TTT to calibration; seven ablations covering signal quality, gating strategy, weight accumulation, loss design, domain ordering, hyperparameter sensitivity, and layer selection confirm that each component is crucial and robust across configurations.
Anonymized Code: \href{https://anonymous.4open.science/r/secl-emnlp26-submission-C890}{anonymous.4open.science/submission-C890}
\end{abstract}

\section{Introduction} \label{sec:introduction}

LLMs are systematically overconfident \cite{jiang_how_2021, xiong_can_2024}, and alignment procedures such as Reinforcement Learning from Human Feedback \citep[RLHF;][]{bai_training_2022} worsen this by rewarding agreement with human preferences over truthfulness \cite{sharma_understanding_2024}. The consequences are practical: in healthcare, where LLMs increasingly support triage and diagnostics \cite{liu_application_2025}, a review of 519 studies found that only 1.2\% measured calibration despite 95.4\% measuring accuracy \cite{bedi_testing_2025}. A model reporting 90\% certainty on questions it answers correctly only 30\% of the time erodes clinician trust and risks patient harm. Addressing this problem needs calibration methods that work without labels and can adapt to new domains at test time.

\begin{figure}[!t]
    \centering
    \includegraphics[width=\columnwidth]{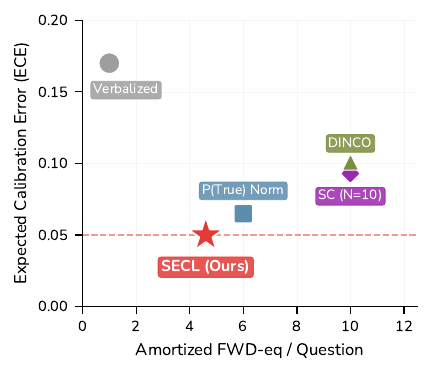}
    \caption{Calibration error vs.\ inference cost for Llama~3.2-3B (lower-left is better). Each point is a calibration method; the $x$-axis measures cost in forward-pass equivalents per question, and the $y$-axis measures Expected Calibration Error (ECE). SECL achieves the lowest calibration error at a fraction of the cost of existing methods.}
    \label{fig:pareto_main}
\end{figure}

\begin{figure*}[!th]
  \centering
  \includegraphics[width=\textwidth]{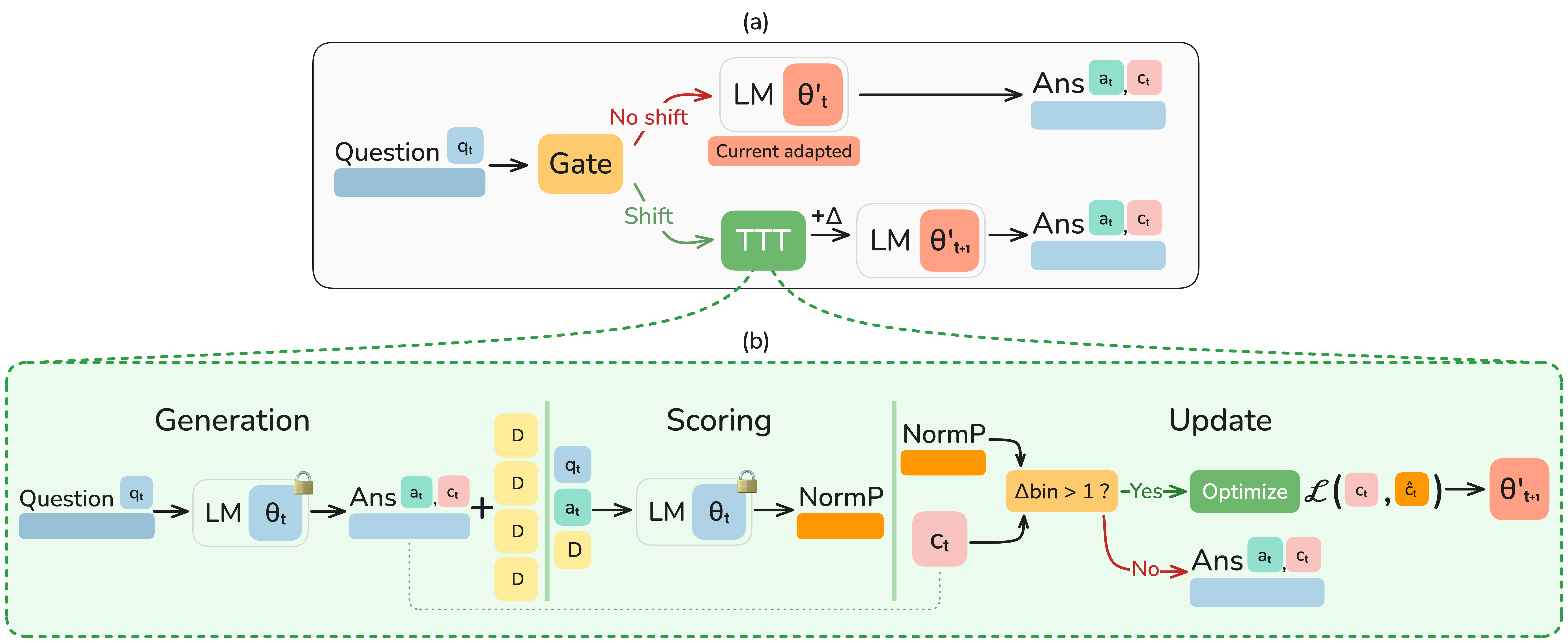}
  \caption{Overview of SECL.
  \textbf{(a)}~\emph{Test-Time Inference.} An entropy-based change detector (Section~\ref{subsec:gating}) monitors the input stream. If no shift is detected, the adapted model $\theta'_t$ is used directly; otherwise, a calibration burst updates it to $\theta'_{t+1}$.
  \textbf{(b)}~\emph{Calibration Burst.} For each of $B{=}50$ questions: the frozen model generates an answer with confidence $c_t$ and distractors, computes $\text{NormP}_{\text{True}}$ (Section~\ref{subsec:signal}), and applies a LoRA update when the two signals disagree by more than one bin (Section~\ref{subsec:ttc}). Weights accumulate across questions without resetting.}
  \label{fig:pipeline}
\end{figure*}

Yet LLMs already contain a better-calibrated signal than the one they verbalize. When asked whether their own answer to a question is correct, LLMs produce probability estimates ($P(\text{True})$) that are substantially better calibrated than the confidence they express during generation \cite{kadavath_language_2022, tian_just_2023}. 
This exposes a systematic gap between \emph{discrimination} and \emph{generation}, with theoretical backing: \citet{kalai_why_2025} shows that a model’s generative error is lower-bounded by roughly twice its discriminative error. Intuitively, a model that cannot reliably produce the correct answer can still often recognize when a given answer, or its own, is wrong. This gap provides a continual source of self-supervision that needs no labeled data.
Prior work documents this phenomenon in LLMs; our experiments test whether the same mechanism can be exploited effectively in small language models.

Prior work on calibration falls into three broad categories, each with a key limitation that SECL addresses (see Section~\ref{sec:related-work} for more details). Sampling-based methods \cite{manakul_selfcheckgpt_2023, kuhn_semantic_2023} measure consistency across multiple generations but are expensive and fail on consistent hallucinations. SECL avoids repeated sampling entirely. Static probing methods \cite{du_haloscope_2024, liu_litcab_2024} analyze internal representations but degrade under distribution shift. SECL adapts continuously via test-time training. Training-based approaches \cite{lin_teaching_2022, stangel_rewarding_2025, damani_binary_2025} can improve calibration within a domain, but many either require supervised labels or degrade OOD under standard Reinforcement Learning (RL) training.
Test-time training (TTT), which adapts model weights to incoming test data instead of relying on a fixed training distribution, originated in computer vision \cite{sun_testtime_2020, wang_tent_2021, wang_continual_2022} and has recently been extended to LLMs for accuracy improvement \cite{hardt_testtime_2024, hu_testtime_2025, sun_learning_2025, zweiger_selfadapting_2025}, but has not yet been applied to calibration due to two obstacles.
First, TTT requires a self-supervision signal at test time; existing calibration signals (sampling-based consistency or discriminative probes) are expensive to compute on every question, making continuous adaptation costly. Second, naive weight updates risk catastrophic forgetting or overfitting to noisy targets, especially when the supervision signal itself is imperfect.

We show that both obstacles can be overcome. SECL uses the generation--discrimination gap as a naturally available scalar target: a normalized $P(\text{True})$ signal computed from the frozen base model serves as self-supervision to adjust verbalized confidence via lightweight Low-Rank Adaptation \citep[LoRA;][]{hu_lora_2022} updates, requiring no labeled data or human supervision (Figure~\ref{fig:pareto_main}). Entropy-based gating triggers adaptation only on distribution shifts, and a conservative directional loss with bounded updates mitigates catastrophic forgetting.
\\Our contributions are as follows:
\begin{itemize}
    \item We introduce the first \textbf{TTT method for calibration}, using the generation--discrimination gap as label-free self-supervision. Entropy-based gating limits adaptation to distribution shifts, so SECL trains only on \textbf{6--26\% of the data} at a lower cost than the signal it distills.
    \item The adapted model \textbf{surpasses} its self-supervision signal and \textbf{matches supervised calibration without labels}, showing that SECL generalizes beyond the training signal.
    \item We provide \textbf{seven ablations} that isolate each design choice, showing that signal quality sets the calibration ceiling and that \textbf{each component is necessary.} SECL is robust across four architectures, forward and reversed domain orderings, and all tested hyperparameters.
\end{itemize}

\section{Related Work} \label{sec:related-work}

\paragraph{Sampling-Based Uncertainty Estimation.}
SelfCheckGPT \cite{manakul_selfcheckgpt_2023} detects hallucinations by comparing each sentence ($S$) against $N{=}20$ sampled passages at the cost of $\mathcal{O}(S \times N)$.
Semantic Entropy \cite{kuhn_semantic_2023} reduces surface sensitivity by clustering semantically equivalent responses before computing entropy, but still requires multiple generations and cannot resolve consistent falsehoods. \citet{ma_semantic_2025} improve detection in single-cluster failure cases by working on penultimate-layer logits. 
\citet{xiong_can_2024} benchmark these and other black-box elicitation methods comprehensively, finding that systematic overconfidence is inherent to all strategies; \citet{heo_llms_2025} confirm this for instruction-tuned models specifically. All sampling methods share two limitations: high cost at inference time, and failure on consistent hallucinations where the model is confidently wrong across all sampled responses \cite{lin_truthfulqa_2021}.

\paragraph{Static Probing and Lightweight Calibration.}
White-box methods bypass sampling by analyzing the model's internals directly. HaloScope \cite{du_haloscope_2024} shows that hallucinations are geometrically distinct in intermediate-to-late layer embeddings, achieving strong AUROC with reasonable cross-dataset transfer, though the authors note degradation under drastic distribution shift. LitCab \cite{liu_litcab_2024} adds a single linear layer (${<}2\%$ of parameters) that predicts a logit bias, reducing ECE by up to 30\%. Although these methods are efficient, they are static: because they are trained offline, they cannot adapt when the input distribution shifts at test time.

\paragraph{Training-Based Calibration.}
\citet{lin_teaching_2022} showed that supervised fine-tuning with calibrated confidence labels produces well-calibrated models, though evaluation did not extend beyond math tasks. \citet{stangel_rewarding_2025} use Reinforcement Learning (RL) with a logarithmic scoring rule to penalize overconfidence. TruthRL \cite{wei_truthrl_2025} uses a ternary reward to incentivize abstention over hallucination when models are uncertain. \citet{damani_binary_2025} use the Brier score \cite{brier_verification_1950} as an RL reward, reducing calibration error by up to 90\% in-domain, but found that standard RL degrades calibration OOD, directly motivating a test-time approach that can adapt to unseen domains.
Prompting methods such as Fact-and-Reflection \cite{zhao_factandreflection_2024} reduce ECE without training but remain static. 
A fundamental limitation of training-based approaches is the difficulty of specifying knowledge boundaries in black-box LLMs: RLHF methods rely on human labels that introduce sycophancy \cite{sharma_understanding_2024}, and RLHF fine-tuning itself degrades calibration \cite{tian_just_2023} despite models retaining well-calibrated internal judgments \cite{kadavath_language_2022}.

\paragraph{The Generation-Discrimination Gap.}
\citet{kadavath_language_2022} established that LLMs' discriminative judgments $P(\text{True})$ are well-calibrated and improve with scale.
\citet{tian_just_2023} extended this to RLHF-tuned models, showing that verbalized confidence is typically better calibrated than conditional token probabilities, even though RLHF degrades the latter.
Building on this line of work, \citet{kalai_why_2025} provided a theoretical basis: generative error is lower-bounded by approximately twice the misclassification rate of the corresponding binary validity problem.
\citet{wang_calibrating_2025} exploit this gap directly: their DINCO method normalizes verbalized confidence across natural language inference (NLI)-reweighted distractors and integrates self-consistency, outperforming prior baselines. However, DINCO is a static inference-time technique that cannot adapt when the underlying truthfulness signal is brittle under distribution shift \cite{haller_llm_2025}. SECL addresses this by distilling the discriminative signal into the model's weights, enabling continuous adaptation. We provide a direct empirical comparison in Section~\ref{subsec:main_results} and Appendix~\ref{app:dinco}.

\paragraph{Test-Time Adaptation for LLMs.}
Test-time training (TTT) adapts model weights using unsupervised signals from incoming test data, originating in computer vision \cite{sun_testtime_2020, wang_tent_2021, wang_continual_2022} and recently extended to LLMs for accuracy improvement \cite{hardt_testtime_2024, hu_testtime_2025, sun_learning_2025, zweiger_selfadapting_2025}. \citet{snell_scaling_2025} showed that adaptive test-time compute allocation can outperform much larger models, and \citet{huang_efficient_2025} use calibrated confidence to allocate such compute, but their goal is accuracy, not calibration. SECL is the first method to apply test-time training to improve \emph{calibration} across domains, leveraging a naturally available supervision signal without any task-specific design.

\section{Methods} \label{sec:methods}
The core idea is simple: when the model encounters a new type of question, it checks whether its stated confidence matches its own self-assessment. If the model claims 90\% confidence but its True/False self-check suggests only 30\%, a small weight update corrects this mismatch. Over time, these corrections accumulate, producing better-calibrated confidence.

Concretely, SECL operates in three stages (Figure~\ref{fig:pipeline}): 
(1) an entropy-based change detector triggers adaptation only on distribution shifts (Section~\ref{subsec:gating}); 
(2) a normalized discriminative signal, $\text{NormP}_{\text{True}}$, scores model answers (Section~\ref{subsec:signal}); 
(3) when this signal disagrees with verbalized confidence, lightweight LoRA updates reduce the gap (Section~\ref{subsec:ttc}). 
Each component is validated individually (Section~\ref{sec:ablation_studies}) and is described below.

\subsection{Adaptive Entropy Gating} \label{subsec:gating}

The calibration procedure (Sections~\ref{subsec:signal}--\ref{subsec:ttc}) requires computing $\text{NormP}_{\text{True}}$ and running LoRA updates for each question. Once the model has adapted to a domain, these updates are redundant; the current LoRA weights already reflect the calibration characteristics of the current distribution. Therefore, calibration is triggered only when the input distribution shifts.

We track the entropy $H_t$ of the model's output token distribution with an exponential moving average (EMA, smoothing factor $\alpha_{\text{ema}}$) and apply the Page-Hinkley (PH) change detection test \cite{page_continuous_1954}. The PH test maintains a cumulative sum:
\begin{equation} \label{eq:ph}
    m_t = \sum_{s=1}^{t} \big(H_s - \bar{H}_t - \epsilon\big),
\end{equation}
where $\bar{H}_t$ is the running mean entropy, $H_s$ the entropy at step $s$, and $\epsilon$ is a tolerance that suppresses false alarms from minor fluctuations. An alarm fires when $m_t - \min_{s \leq t} m_s > \lambda$; a higher $\lambda$ requires a larger cumulative entropy deviation before triggering, controlling the trade-off between responsiveness and false alarm rate (sensitivity analysis in Section~\ref{sec:ablation_studies}). Upon detection, cumulative statistics are reset, and a calibration burst of $B$ consecutive questions is initiated. 
Processing multiple questions per burst is crucial: a single LoRA update provides too little signal for stable adaptation, whereas a burst amortizes the cost of entering calibration mode and allows corrections to accumulate across diverse questions from the new distribution.

LoRA weights accumulate across domains without resetting. When a new distribution shift is detected, the subsequent calibration burst adapts the existing accumulated weights rather than starting from scratch, allowing calibration knowledge from earlier domains to compound. Datasets and domain ordering are described in Section~\ref{subsec:setup}, with ordering sensitivity analyzed in Section~\ref{sec:ablation_studies}.

\subsection[Normalized P(True) as Self-Supervision]{Normalized $P(\text{True})$ as Self-Supervision} \label{subsec:signal}
Given a question $q$ and the model's generated answer $a$, we compute $P_{\text{True}}(a \mid q)$: the token probability of ``True'' when the model is asked ``Is the following answer to the question correct? (True/False)''. Following \citet{kadavath_language_2022}, this discriminative signal is well-calibrated and improves with scale, making it a stronger supervision target than the model's own verbalized confidence $c$.

Following \citet{lin_teaching_2022} and \citet{tian_just_2023}, we elicit verbalized confidence by prompting the model to state a confidence bin (0--9) alongside its answer, where each bin corresponds to a 10-percentage-point interval (e.g., bin~7 $\approx$ 70--80\%). We use 10 bins to ensure confidence can be expressed as a single generated token \cite{naeini_obtaining_2015, guo_calibration_2017}. Rather than taking the argmax bin (hard readout), we compute a \emph{soft} confidence as the expected value over the digit-token probability distribution:
\begin{equation} \label{eq:soft_conf}
    c = \sum_{k=0}^{9} P(\text{bin}_k) \cdot 
    \frac{k + 0.5}{10},
\end{equation}
where $P(\text{bin}_k)$ is the model's output probability for digit token $k$. This soft readout preserves information from the full distribution and provides a differentiable signal for the mean squared error (MSE) loss in Section~\ref{subsec:ttc}.

Raw $P_{\text{True}}$ suffers from suggestibility bias, the model tends to affirm any answer presented to it, inflating $P_{\text{True}}$ regardless of correctness \cite{wang_calibrating_2025}, and degrades under distribution shift \cite{haller_llm_2025}. We address both by normalizing $P_{\text{True}}$ across distractor answers. For multiple-choice questions, we use the given answer options as distractors. For open-ended questions, we generate $K{=}4$ plausible alternatives. The normalized signal is:
\begin{equation} \label{eq:normp}
    \text{NormP}_{\text{True}}(a) = \frac{e_a}{e_a + \sum_{k=1}^{K} e_{d_k}},
\end{equation}
where $e_x = \exp\!\big(P_{\text{True}}(x) / \tau\big)$, with $\tau$ as a model-specific temperature (see Section~\ref{subsec:setup}); the choice of $\tau$ is discussed in Section~\ref{sec:ablation_studies}.
This softmax over distractors converts the raw signal into a \emph{relative} confidence that accounts for baseline suggestibility.

Unlike DINCO, we use a simple softmax over distractors without NLI reweighting or self-consistency. The resulting $\text{NormP}_{\text{True}}$ provides a continuous training target, which is discretized into the same 10 equal-width bins used for verbalized confidence (Eq.~\ref{eq:soft_conf}), ensuring a common scale between supervision target and model output.

\subsection{Test-Time Calibration via LoRA} \label{subsec:ttc}
When the model's verbalized confidence disagrees with $\text{NormP}_{\text{True}}$, we update the model to reduce this disagreement. Updates are applied via LoRA \cite{hu_lora_2022} on intermediate-to-late transformer layers, motivated by the finding that calibration-relevant representations concentrate in these layers \cite{du_haloscope_2024}. Architecture-specific layer configurations are reported in Appendix~\ref{app:layer_ablation}.

\paragraph{Training target.}
Although distractor normalization suppresses systematic biases (Section~\ref{subsec:signal}), the discriminative signal can still be noisy on individual questions. We therefore do not use $\text{NormP}_{\text{True}}$ directly as the training target. Instead, rather than jumping directly to the discriminative estimate, we nudge confidence toward it in small, bounded steps.

Let $c_i$ be the model's verbalized confidence and $c_i^*$ the $\text{NormP}_{\text{True}}$ value for question $i$.
The training target is:
\begin{equation} \label{eq:target}
    \hat{c}_i = c_i + \alpha_{\text{step}} \cdot \operatorname{clip}\!\big(c_i^* - c_i, \; {-}\delta, \; \delta\big),
\end{equation}
where $\operatorname{clip}(x, -\delta, \delta) = \max(-\delta, \min(x, \delta))$ clamps the correction magnitude to the interval $[-\delta, \delta]$, $\alpha_{\text{step}}$ controls the correction step size, and $\delta$ caps the maximum single-step adjustment.
We fix $\alpha_{\text{step}}$ and $\delta$ across all models (values in Section~\ref{subsec:setup}).

\paragraph{Loss and optimization.}
The training loss is mean squared error between verbalized confidence and the directional target:
\begin{equation} \label{eq:loss}
    \mathcal{L}_i = \big(c_i - \hat{c}_i\big)^2.
\end{equation}
We optimize with AdamW \cite{loshchilov_decoupled_2019}; learning rate and epoch count are reported in Section~\ref{subsec:setup}.
Crucially, $\text{NormP}_{\text{True}}$ is  computed from the \emph{base model without LoRA adapters}, ensuring the supervision signal is not corrupted by ongoing adaptation.

\paragraph{Bin-gate filter.}
Not every question requires calibration.
We skip training when the model is already approximately calibrated: specifically, when $|\text{bin}(c_i) - \text{bin}(c_i^*)| \leq 1$.
This avoids gradient updates on questions where verbalized and discriminative confidence already agree, reducing computation and limiting noise from marginal disagreements.
We use threshold~1 as default; the effect on training fraction and ECE is shown in Section~\ref{sec:ablation_studies}.

\section{Experiments} \label{sec:experiments}

\begin{figure*}[t]
  \centering
  \begin{minipage}[t]{0.48\textwidth}
      \centering
      \includegraphics[width=\textwidth]{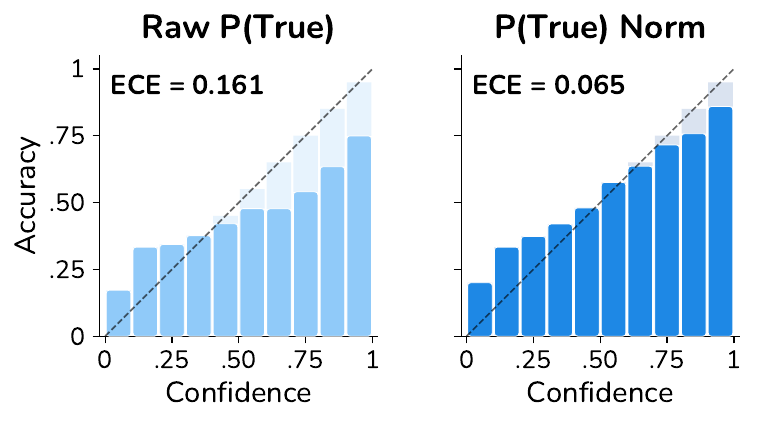}
      \caption{Reliability diagrams for Llama~3.2-3B (2{,}000 questions). \textbf{Left:}~Raw P(True) is poorly calibrated (ECE\,=\,0.161). \textbf{Right:}~After temperature normalization, P(True) Norm is substantially better calibrated (ECE\,=\,0.065), confirming the value of the discriminative signal.}
      \label{fig:gap}
  \end{minipage}
  \hfill
  \begin{minipage}[t]{0.48\textwidth}
      \centering
      \includegraphics[width=\textwidth]{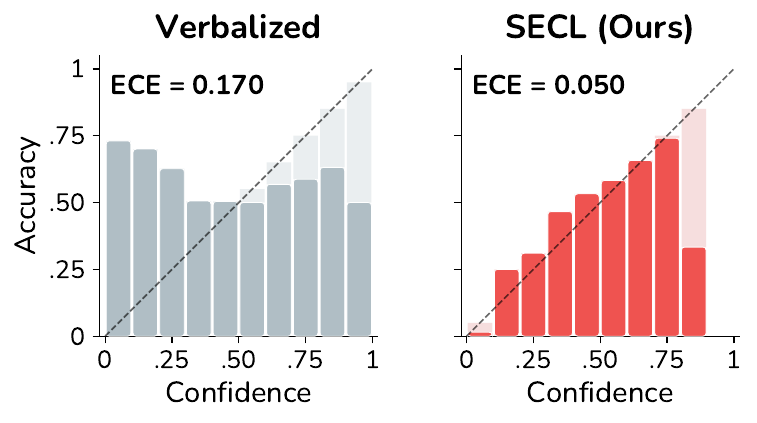}
      \caption{Reliability diagrams for Llama~3.2-3B. \textbf{Left:}~Verbalized baseline (ECE\,=\,0.170). \textbf{Right:}~After SECL (ECE\,=\,0.050, 71\,\% reduction), surpassing its own training signal (Figure~\ref{fig:gap}). The empty top bin shows that SECL eliminates the baseline's false near-100\% certainty.}
      \label{fig:reliability}
  \end{minipage}
\end{figure*}

\subsection{Setup} \label{subsec:setup}

\paragraph{Datasets.}
We evaluate on four datasets:
\textbf{GSM8K} \cite{cobbe_training_2021}, grade-school math word problems (train split, first 500);
\textbf{MMLU} \cite{hendrycks_measuring_2021}, multiple-choice knowledge questions across 57 subjects (test split, 500 sampled round-robin across subjects);
\textbf{ARC Challenge} \cite{clark_think_2018}, grade-school science questions (test split, first 500);
and \textbf{TruthfulQA} \cite{lin_truthfulqa_2021}, adversarially constructed misconception questions (validation split, MC1 variant, first 500).
These four domains differ in reasoning type, difficulty, and degree of model overconfidence.

\paragraph{Continual protocol.}
The four domains are presented sequentially (GSM8K $\to$ MMLU $\to$ ARC $\to$ TruthfulQA), forming a single stream of 2{,}000 questions with distribution shifts between domains. Per-domain and ordering results are in Appendix~\ref{app:per_domain} and Appendix~\ref{app:domain_order}; an additional open-ended variant using TruthfulQA generation answers is reported in Appendix~\ref{app:tqgen}.

\paragraph{Models.}
We evaluate four instruction-tuned small language models spanning different architectures: Llama~3.2-3B and Llama~3.1-8B \cite{grattafiori_llama_2024}, Gemma~2-2B \cite{team_gemma_2024}, and Phi~3.5-Mini (3.8B) \cite{abdin_phi3_2024}. These were selected because all four exhibit a measurable generation--discrimination gap: their $\text{NormP}_{\text{True}}$ signal is better calibrated than their verbalized confidence, which is the prerequisite for SECL.

\paragraph{Baselines.}
We compare against two baselines representing the cost–quality extremes for label-free calibration.
\emph{Verbalized}: the model's verbalized confidence $c$ (soft readout, Eq.~\ref{eq:soft_conf}), used directly with no adaptation, representing a zero-cost baseline.
\emph{P(True) Norm}: the distractor-normalized discriminative signal (Eq.~\ref{eq:normp}) reported directly as the confidence estimate. This serves as an upper-bound reference for signal quality; it requires five discriminative forward passes per question, does not modify model weights, and represents the quality of the supervision signal that SECL distills.

\paragraph{Metrics.}
We report Expected Calibration Error (ECE; \citealt{naeini_obtaining_2015, guo_calibration_2017}) as our primary calibration metric, Brier score \cite{brier_verification_1950} as a composite measure of calibration and discrimination, AUROC for discrimination quality, and task accuracy to verify that calibration updates do not degrade correctness. Formal definitions are provided in Appendix~\ref{app:metrics}.

\paragraph{Implementation details.}
All values below are the best configuration determined through the ablation studies in Section~\ref{sec:ablation_studies}, Appendix~\ref{app:hp_sensitivity}, and Appendix~\ref{app:layer_ablation}.
LoRA (rank $r{=}8$, $\alpha{=}16$) is applied to the query and value projection matrices of the last 4--8 transformer layers, targeting intermediate-to-late layers where calibration-relevant representations concentrate \cite{du_haloscope_2024}. The exact layer range is model-dependent (Appendix~\ref{app:layer_ablation}). This modifies 328K--786K parameters (${\sim}$0.01--0.02\% of total) per model.
We optimize with AdamW (learning rate $5 \times 10^{-5}$) for 3 epochs per question. The directional target uses $\alpha_{\text{step}} = 0.5$ and $\delta = 0.15$; both are robust across tested ranges (Appendix~\ref{app:hp_sensitivity}).
The entropy gate uses $\alpha_{\text{ema}} = 0.05$, $\epsilon = 0.05$, detection threshold $\lambda = 3.0$ (sensitivity analysis in Section~\ref{sec:ablation_studies}), burst length $B = 50$, and a warmup of 30 questions before first detection.
The normalization temperature $\tau$ is fixed per model family from preliminary exploratory runs of the $P(\text{True})$ baseline (Section~\ref{sec:ablation_studies}) and then held constant across datasets: $\tau{=}0.7$ for Llama~3.2-3B, $\tau{=}1.5$ for Gemma and Phi, and $\tau{=}3.0$ for Llama~3.1-8B.
All experiments run on NVIDIA A100 (80\,GB) and RTX A6000 (48\,GB) GPUs.
Hyperparameters are in Appendix~\ref{app:hyperparams}. The main table reports fixed-seed runs; Llama multi-seed results appear in Appendix~\ref{app:llama_multiseed}.

\subsection{Main Results} \label{subsec:main_results}

\begin{table*}[t]
\centering
\renewcommand{\arraystretch}{1.35}
\resizebox{\textwidth}{!}{%
\setlength{\tabcolsep}{3.5pt}
\begin{tabular}{@{}l@{\hspace{4pt}}c*{12}{c}@{}}
\toprule
& & \multicolumn{3}{c}{\textbf{Llama 3.2-3B}}
& \multicolumn{3}{c}{\textbf{Llama 3.1-8B}}
& \multicolumn{3}{c}{\textbf{Gemma 2-2B}}
& \multicolumn{3}{c}{\textbf{Phi 3.5-Mini}} \\
\cmidrule(l{2pt}r{2pt}){3-5} \cmidrule(l{2pt}r{2pt}){6-8} \cmidrule(l{2pt}r{2pt}){9-11} \cmidrule(l{2pt}r{2pt}){12-14}
\textbf{Method} & \textbf{Cost}{\scriptsize$\downarrow$}
& ECE{\scriptsize$\downarrow$} & Brier{\scriptsize$\downarrow$} & AUROC{\scriptsize$\uparrow$}
& ECE{\scriptsize$\downarrow$} & Brier{\scriptsize$\downarrow$} & AUROC{\scriptsize$\uparrow$}
& ECE{\scriptsize$\downarrow$} & Brier{\scriptsize$\downarrow$} & AUROC{\scriptsize$\uparrow$}
& ECE{\scriptsize$\downarrow$} & Brier{\scriptsize$\downarrow$} & AUROC{\scriptsize$\uparrow$} \\
\midrule
Verbalized & \textbf{1}
& .170 & .292 & .510
& .225 & .258 & .684
& .256 & .314 & .558
& .251 & .275 & .600 \\

P(True) Norm & 6
& .065 & .223 & .694
& .120 & .211 & .718
& .141 & .259 & \textbf{.650}
& .154 & .227 & .675 \\

DINCO$^\dagger$ & {$\sim$}10
& .101 & \textbf{.207} & \textbf{.762}
& .117 & \textbf{.210} & \textbf{.756}
& .408 & .410 & .566
& \textbf{.110} & \textbf{.212} & \textbf{.749} \\

\textbf{SECL (Ours)} & 1.8--4.6
& \textbf{.050} & .241 & .587
& \textbf{.083} & .222 & .643
& \textbf{.056} & \textbf{.254} & .548
& \textbf{.110} & .251 & .521 \\
\bottomrule
\end{tabular}}%
\caption{Overall results across the full 2{,}000-question stream. Best values per column in \textbf{bold} (lowest for ECE/Brier/Cost, highest for AUROC). Cost is amortized forward-pass equivalents per question; SECL's range reflects model-dependent gating rates (Appendix~\ref{app:cost}). Accuracy is preserved within 1 percentage point for all models (Appendix~\ref{app:main_results}). $^\dagger$\citet{wang_calibrating_2025}.}
\label{tab:main_results}
\end{table*}

\paragraph{SECL improves calibration across all four models.}
Table~\ref{tab:main_results} reports overall metrics across the full 2{,}000-question stream. Compared with the verbalized baseline, ECE reductions range from 56\% on Phi to 78\% on Gemma. SECL also improves over its own supervision signal, P(True) Norm, on all four models, despite training on only 6--26\% of the stream (Table~\ref{tab:gating_ablation}). 
SECL surpasses its own supervision signal, showing that the model internalizes the discriminative signal well enough to generalize beyond the questions it was trained on.

\paragraph{Largest gains appear where miscalibration is most severe.}
MMLU and TruthfulQA show the largest drop; GSM8K and ARC show smaller gains. Per-domain breakdowns appear in Appendix~\ref{app:per_domain}.

\paragraph{Comparison with supervised post-hoc calibration.}
Supervised baselines (temperature \& Platt scaling \cite{platt_probabilistic_1999} fitted with ground-truth labels via 5-fold cross-validation achieve lower ECE but at the cost of severe confidence compression, collapsing predictions into narrow ranges (e.g., [0.44,\,0.56] for Llama). SECL preserves a wide confidence range without any labels (Appendix~\ref{app:posthoc}).

\paragraph{Accuracy is preserved.}
Accuracy differs by $<1$ percentage point overall (Table~\ref{tab:main_results}); per-domain shifts are at most 3 percentage points. Unlike RL-based methods that can harm task performance \cite{damani_binary_2025, stangel_rewarding_2025}, SECL updates only the confidence representation, not the model's task behavior.

\paragraph{Computational cost.}
SECL is cheaper than P(True) Norm and 2--5$\times$ cheaper than DINCO. Entropy gating is key: models that trigger infrequently (Gemma: 5.9\%, Phi: 8.0\%) reduce cost to less than a third of the P(True) Norm baseline (Appendix~\ref{app:cost}).

SECL reduces ECE under both forward and reversed orderings on all four models, with reductions of 16--71\%. Per-domain ECE wins shift with ordering (e.g., reversed Llama~3B reaches 0.032 on MMLU vs.\ 0.067 forward), indicating adaptation to whichever domain the model encounters.

\paragraph{Comparison with DINCO.}
SECL also outperforms DINCO \cite{wang_calibrating_2025}, the closest inference-time calibration method, which combines NLI-weighted distractor normalization with self-consistency sampling at ${\sim}$10 forward passes per question. SECL achieves lower or equal ECE on all four models at 2--5$\times$ lower amortized cost. On Gemma, DINCO fails entirely: its ECE is worse than the unadapted baseline, and accuracy drops substantially, confirming that NLI-based reweighting is not robust across architectures. DINCO achieves stronger discrimination on Llama and Phi through multiple sampling passes, though its default beam-search answer selection yields lower task accuracy than greedy decoding. Figure~\ref{fig:pareto_main} visualizes the cost--calibration trade-off; details in Appendix~\ref{app:dinco}.

\subsection{Ablation Studies} \label{sec:ablation_studies}

We validate every design decision via seven targeted ablations, each isolating one component while keeping all others fixed. Results are on Llama~3.2-3B unless otherwise noted; domain ordering ablations cover all four models. Together, these ablations establish three key insights: (1)~the adaptation mechanism is cheap, requiring only 25.6\% of the stream; (2)~signal quality determines the ceiling of calibration improvement; and (3)~each component is individually necessary.

\paragraph{Entropy gating matches always-on adaptation.}
Table~\ref{tab:gating_ablation} compares four gating strategies on the full 2{,}000-question stream.

Entropy gating determines \emph{when} to enter a calibration burst; the bin-gate filters \emph{which} questions within the burst receive updates. The combined entropy gate + bin-gate processes only 25.6\% of the stream while matching always-on ECE (0.050 vs.\ 0.047): a 4$\times$ reduction in calibration compute with no measurable loss in calibration quality. This makes SECL cheaper per question than the raw P(True) Norm baseline it learns from.

\paragraph{Directional loss improves over plain MSE.}
Replacing the directional target with plain MSE toward $\text{NormP}_{\text{True}}$ degrades ECE from 0.052 to 0.085 (Appendix~\ref{app:hp_sensitivity}). Conservative clipping prevents overshooting noisy targets, confirming that the update direction matters more than its magnitude.

\paragraph{Calibration knowledge must accumulate across questions.}
Resetting LoRA weights after each question yields ECE 0.237 and degrades AUROC to 0.484, \emph{below chance}, indicating that isolated single-question updates destroy the model's confidence signal entirely (Table~\ref{tab:ablation_accum}). Weight accumulation achieves ECE 0.050: a 4.7$\times$ improvement, demonstrating that calibration knowledge compounds across questions and that the continual learning design is essential, not optional.

\begin{table}[!t]
\centering
\small
\renewcommand{\arraystretch}{1.35}
\begin{tabular}{@{}lccc@{}}
\toprule
\textbf{Model}
  & \textbf{Verbalized} & \textbf{SECL}~$\mathcal{O}$
  & \textbf{SECL}~$\mathcal{R}$ \\
\midrule
Llama 3.2-3B  & .170 & \textbf{.050}\,{\scriptsize($-$71\%)} & .068\,{\scriptsize($-$60\%)} \\
Llama 3.1-8B  & .225 & \textbf{.083}\,{\scriptsize($-$63\%)} & .189\,{\scriptsize($-$16\%)} \\
Gemma 2-2B    & .256 & \textbf{.056}\,{\scriptsize($-$78\%)} & .149\,{\scriptsize($-$42\%)} \\
Phi 3.5-Mini  & .251 & \textbf{.110}\,{\scriptsize($-$56\%)} & .173\,{\scriptsize($-$31\%)} \\
\bottomrule
\end{tabular}
\caption{Domain order robustness (ECE).
$\mathcal{O}$: GSM8K$\to$MMLU$\to$ARC$\to$TruthfulQA (default).
$\mathcal{R}$: reversed.
SECL improves over the verbalized baseline under both orderings
for all models; per-domain breakdowns in Appendix~\ref{app:domain_order}.}
\label{tab:domain_order}
\end{table}

\paragraph{Distractor normalization is essential for signal quality.}
Raw $P(\text{True})$ already improves over the verbalized baseline (ECE: 0.161 vs.\ 0.170), confirming the generation--discrimination gap. Distractor normalization reduces ECE further to 0.065, a 60\% improvement, by converting absolute affirmation into relative preference among candidates and suppressing the suggestibility bias identified by \citet{wang_calibrating_2025} (Appendix~\ref{app:ptrue_norm}).

\paragraph{Signal quality determines the calibration ceiling.}
This is the most revealing ablation. Replacing $\text{NormP}_{\text{True}}$ with Self-Consistency \cite{wang_selfconsistency_2023} ($N{=}10$) as SECL's training target, while keeping LoRA adaptation, entropy gating, and all hyperparameters identical, degrades ECE from 0.050 to 0.432, which is 2.5$\times$ \emph{worse} than the untrained baseline (0.170). The adaptation mechanism faithfully distills whatever signal it receives: Self-Consistency's systematic overconfidence propagates directly into verbalized confidence, while $\text{NormP}_{\text{True}}$'s tighter correspondence to correctness probability yields well-calibrated outputs (Appendix~\ref{app:target_ablation}).

\paragraph{Calibration is robust to domain ordering.}
Table~\ref{tab:domain_order} confirms that SECL improves over the baseline under both forward and reversed orderings on all four models, with ECE reductions of 16--71\%.

\paragraph{Hyperparameter sensitivity.}
Appendix~\ref{app:hp_sensitivity} reports sensitivity to each hyperparameter on Llama~3.2-3B. SECL is robust to most hyperparameters: $\alpha_{\text{step}}$ and $\delta$ have minimal impact on ECE (0.064--0.067 across tested values). The single consequential choice is burst length $B$: $B = 20$ yields ECE 0.114 versus 0.050 for $B = 50$, confirming that sufficient distillation per trigger is critical. This robustness simplifies deployment: only one hyperparameter requires careful tuning.

\begin{table}[!t]
\centering
\small
\setlength{\tabcolsep}{4pt}
\renewcommand{\arraystretch}{1.35}
\resizebox{\columnwidth}{!}{%
\begin{tabular}{@{}lcccc@{}}
\toprule
\textbf{Strategy} & \textbf{ECE}$\downarrow$ 
& \textbf{Brier}$\downarrow$ & \textbf{AUROC}$\uparrow$ 
& \textbf{Trained} \\
\midrule
Always-on MSE        & 0.047 & 0.240 & 0.593 & 100\%  \\
Bin-gate ($\leq\!1$)    & 0.052 & 0.242 & 0.585 & 55.8\% \\
Bin-gate ($\leq\!2$)    & 0.044 & 0.242 & 0.578 & 32.6\% \\
Entropy-gated ($B\!=\!50$) & 0.050 & 0.241 & 0.587 & 25.6\% \\
\bottomrule
\end{tabular}
}
\caption{Gating strategy ablation (Llama~3.2-3B). Entropy-gated 
bursts achieve comparable ECE to always-on training while 
requiring only 25.6\% of questions.}
\label{tab:gating_ablation}
\vspace{-1em}
\end{table}

\section{Conclusion} \label{sec:conclusion}

We introduced SECL, a test-time training pipeline that exploits the generation--discrimination gap as label-free self-supervision to continuously improve calibration without labeled data or human supervision. Across four small language models from three model families and four diverse domains, SECL reduces ECE by 56--78\% while preserving task accuracy, training on only 6--26\% of the question stream via entropy-gated adaptation. We conducted seven ablation studies to explore our pipeline and demonstrate that the method is robust to variations in domain ordering and hyperparameter settings, while its success comes from high-quality signal distillation, the necessity of weight accumulation, and entropy-gated adaptation.

Our results suggest a broader principle: when a model's ability to evaluate exceeds its ability to generate, the gap can be distilled back into the model's outputs via our self-supervision pipeline. Calibration is a natural first target because the discriminative signal is scalar and cheap to compute, but the same approach applies to factual accuracy or reasoning consistency wherever an analogous evaluation--generation gap exists. This gap \cite{kalai_why_2025} is not a deficiency to be tolerated but a resource to be exploited, one that is theoretically expected to grow with scale. Scaling SECL to larger models and integrating richer discriminative signals are the direct next steps.

\clearpage
\section*{Limitations} \label{sec:limitations}

\paragraph{Signal quality bounds improvement.}
SECL's calibration gains are bounded by the quality of the NormP$_{\text{True}}$ supervision signal. When the discriminative signal is only marginally better than verbalized confidence, distillation yields smaller improvements. On Gemma TruthfulQA, for example, SECL reduces ECE from 0.267 (Verbalized) to 0.210, but P(True) Norm achieves 0.104, indicating that the distillation captures only part of the available signal on adversarially constructed questions. Integrating richer discriminative signals beyond binary P(True), such as multi-step verification or ensemble-based judgments, could close this gap. SECL requires a measurable generation--discrimination gap, which we verified for all four evaluated model families. For models where this gap is absent, such as Qwen~2.5-3B \cite{qwen_qwen25_2025} where NormP$_{\text{True}}$ underperforms the verbalized baseline at all tested temperatures (Appendix~\ref{app:qwen}), SECL correctly produces no improvement, since no useful signal exists to distill. Characterizing when and why this gap closes is an important direction for future work.

\paragraph{Per-domain calibration is not uniformly improved.}
ARC ECE increases slightly in the forward ordering across all three smaller models (e.g., Llama: 0.095 $\to$ 0.112), but this effect reverses under alternative orderings (Appendix~\ref{app:domain_order}), indicating a domain-sequencing artifact from cumulative LoRA weight transfer rather than a systematic limitation. Aggregate ECE remains improved under every ordering tested.

\paragraph{Calibration--discrimination trade-off.}
SECL improves AUROC on Llama~3.2-3B (+0.077) but degrades it on Phi ($-$0.079) and on Llama~3.1-8B ($-$0.041; Appendix~\ref{app:scaling_8b}). Since the LoRA updates target the confidence token, they can redistribute probability mass in ways that improve bin-level calibration at the cost of per-question ranking quality. Brier score, which captures both components, improves across all models, but applications requiring fine-grained discrimination should weigh this trade-off.

\paragraph{Hyperparameter sensitivity.}
The burst size $B$ is the most important hyperparameter: $B{=}20$ yields ECE 0.114 versus 0.050 for $B{=}50$ on Llama (Section~\ref{sec:ablation_studies}), indicating that sufficient distillation per trigger is necessary. In deployment settings with very rapid distribution shifts (fewer than 50 questions per domain), the method may not accumulate enough training signal.

\paragraph{Scale.}
We evaluate models up to 8B parameters. While the generation--discrimination gap is theoretically expected to widen with scale \cite{kalai_why_2025}, we have not verified SECL's effectiveness beyond 8B, where the computational cost of LoRA updates would also increase.

\section*{Ethics and Societal Impact} \label{sec:ethics}

SECL reduces ECE by 56--78\% without human labels or held-out data. This directly lowers the barrier to calibrated deployment: only 1.2\% of 519 healthcare LLM studies measured calibration \cite{bedi_testing_2025}, partly because existing methods require labeled validation sets that are expensive to curate. By operating entirely at test time with self-generated supervision, SECL makes calibration feasible in settings where annotation budgets are limited or domain expertise is scarce.

Our target ablation (Section~\ref{sec:ablation_studies}) carries an important ethical lesson: substituting Self-Consistency as the supervision signal makes calibration 2.5$\times$ worse than the unadapted baseline. This demonstrates that self-supervised calibration methods can \emph{amplify} overconfidence when the supervision signal is poorly chosen. Practitioners deploying any self-supervised calibration approach, including SECL, should verify that the underlying signal genuinely correlates with correctness before trusting the resulting confidence scores.

Two further risks merit attention. First, calibration does not imply correctness: a well-calibrated model reporting 60\% confidence is still wrong 40\% of the time. Our AUROC results (Section~\ref{subsec:main_results}) show that SECL can degrade per-question ranking quality on some models while improving bin-level calibration, meaning users should not assume that higher confidence always identifies more reliable answers. Second, SECL adapts continuously at test time, so its confidence characteristics evolve with the input stream. This complicates auditing and reproducibility compared to static calibration methods; logging the gating decisions and LoRA checkpoints would be necessary for post-hoc review in regulated domains.

All experiments use public benchmarks (GSM8K, MMLU, ARC-Challenge, TruthfulQA) with standard splits. No human data was collected.

\newpage
\bibliography{26_emnlp_selfcalibrating}

\clearpage
\appendix
\renewcommand{\arraystretch}{1.15}

\section{Experimental Setup Details} \label{app:setup}

\paragraph{Datasets and splits.}
We evaluate on four domains presented sequentially: 
GSM8K \citep{cobbe_training_2021} (\texttt{train} split), 
MMLU \citep{hendrycks_measuring_2021} (\texttt{test} split), 
ARC-Challenge \citep{clark_think_2018} (\texttt{test} split), 
and TruthfulQA \citep{lin_truthfulqa_2021} (\texttt{validation} split, MC1 variant).
We sample 500 questions per domain for a total of 2{,}000 questions per run.
The default domain order is GSM8K $\to$ MMLU $\to$ ARC $\to$ TruthfulQA.

\paragraph{Models.}
We evaluate four instruction-tuned small language models spanning 2--8B parameters:
Llama~3.2-3B-Instruct and Llama~3.1-8B-Instruct \citep{grattafiori_llama_2024},
Gemma~2-2B-IT \citep{team_gemma_2024},
and Phi~3.5-Mini-Instruct (3.8B) \citep{abdin_phi3_2024}.

\paragraph{Trainable parameters.}
Table~\ref{tab:lora_params} reports the exact LoRA parameter counts per model.
The variation reflects architectural differences (Phi uses a fused \texttt{qkv\_proj} 
module rather than separate \texttt{q\_proj}/\texttt{v\_proj}).

\begin{table}[htbp]
\centering
\small
\resizebox{\columnwidth}{!}{%
\begin{tabular}{lcccc}
\toprule
\textbf{Model} & \textbf{Total params} & \textbf{LoRA layers} & \textbf{LoRA params} & \textbf{\% of total} \\
\midrule
Llama 3.2-3B  & 3.21B & 4  & 327{,}680 & 0.010\% \\
Llama 3.1-8B  & 8.03B & 8  & 1{,}048{,}576 & 0.013\% \\
Gemma 2-2B    & 2.61B & 8  & 491{,}520 & 0.019\% \\
Phi 3.5-Mini  & 3.82B & 8  & 786{,}432 & 0.021\% \\
\bottomrule
\end{tabular}}%
\caption{LoRA trainable parameters per model (rank $r{=}8$).
Llama~3.2-3B: last 4 layers (24--27 of 28);
Llama~3.1-8B: last 8 layers;
Gemma and Phi: last 8 layers
(Gemma: 18--25 of 26; Phi: 24--31 of 32).}
\label{tab:lora_params}
\end{table}

\paragraph{Compute and software.}
All experiments were conducted on a shared workstation equipped with NVIDIA A100 (80\,GB) and RTX A6000 (48\,GB) GPUs.
Each 2{,}000-question SECL run takes approximately 2--4 GPU-hours on a single GPU, depending on model size.
We use PyTorch~2.9 \citep{paszke_pytorch_2019}, HuggingFace Transformers~4.57 \citep{wolf_transformers_2020}, and the PEFT library (v0.17) with CUDA~12.8.

\section{Metric Definitions} \label{app:metrics}

\emph{Expected Calibration Error} (ECE; \citealt{naeini_obtaining_2015, guo_calibration_2017}; lower is better). Predictions are divided into $M{=}10$ equal-width bins $B_1, \dots, B_M$ by confidence value:
\begin{equation}
    \text{ECE} = \sum_{m=1}^{M} \frac{|B_m|}{N} 
    \,\big|\text{acc}(B_m) - \text{conf}(B_m)\big|,
\end{equation}
where $N$ is the total number of predictions, $\text{acc}(B_m)$ is the fraction of correct predictions in bin $B_m$, and $\text{conf}(B_m)$ is the mean predicted confidence. A perfectly calibrated model achieves $\text{ECE} = 0$.

\emph{Brier score} \cite{brier_verification_1950} (lower is better). The mean squared error between predicted confidence and binary correctness:
\begin{equation}
    \text{BS} = \frac{1}{N} \sum_{i=1}^{N} 
    (c_i - y_i)^2,
\end{equation}
where $c_i \in [0,1]$ is the predicted confidence and $y_i \in \{0,1\}$ indicates correctness. The Brier score decomposes into calibration and refinement components.

\emph{AUROC} (area under the ROC curve; higher is better). The area under the receiver operating characteristic curve, treating predicted confidence as a score for binary classification of correctness. AUROC measures discrimination, whether higher confidence corresponds to higher correctness, independent of calibration.

\emph{Adaptive ECE} (AdaECE). Same as ECE but with equal-mass bins (lower is better); we report it to verify robustness to binning choice.

\emph{Accuracy}. Fraction of correct answers; we report it to verify that calibration updates do not degrade task performance.

\section{Hyperparameter Settings} \label{app:hyperparams}

\begin{table}[H]
\centering
\small
\resizebox{\columnwidth}{!}{%
\begin{tabular}{llc}
\toprule
\textbf{Component} & \textbf{Parameter} & \textbf{Value} \\
\midrule
\multirow{6}{*}{LoRA}
  & Rank $r$ & 8 \\
  & Scaling factor $\alpha$ & 16 \\
  & Target layers (Llama 3.2-3B) & Last 4 \\
  & Target layers (Llama 3.1-8B, Gemma, Phi) & Last 8 \\
  & Target modules (Llama, Gemma) & \texttt{q\_proj}, \texttt{v\_proj} \\
  & Target modules (Phi) & \texttt{qkv\_proj} \\
\midrule
\multirow{3}{*}{Optimization}
  & Optimizer & AdamW \\
  & Learning rate & $5 \times 10^{-5}$ \\
  & Epochs per question & 3 \\
\midrule
\multirow{2}{*}{Directional loss}
  & Step size $\alpha_{\text{step}}$ & 0.5 \\
  & Clip bound $\delta$ & 0.15 \\
\midrule
\multirow{1}{*}{Bin-gate}
  & Threshold & 1 bin \\
\midrule
\multirow{4}{*}{Page-Hinkley}
  & Tolerance $\epsilon$ & 0.05 \\
  & Detection threshold $\lambda$ & 3.0 \\
  & EMA smoothing $\alpha_{\text{ema}}$ & 0.05 \\
  & Warmup period & 30 questions \\
\midrule
\multirow{1}{*}{Burst}
  & Burst size $B$ & 50 \\
\midrule
\multirow{4}{*}{Normalization $\tau$}
  & Llama 3.2-3B & \textbf{0.7} \\
  & Gemma 2-2B & \textbf{1.5} \\
  & Phi 3.5-Mini (3.8B) & \textbf{1.5} \\
  & Llama 3.1-8B & \textbf{3.0} \\
\midrule
\multirow{2}{*}{Other}
  & Weight accumulation & On \\
  & Distractors $k$ & 4 \\
\bottomrule
\end{tabular}}%
\caption{Full hyperparameter settings used across all experiments.
Model-specific values are noted where applicable.}
\label{tab:hyperparams}
\end{table}

\section{Main Results} \label{app:main_results}

Table~\ref{tab:main_results_full} reports overall metrics for all methods and all four models.
We report both standard ECE (10 equal-width bins) and Adaptive ECE (AdaECE; 10 equal-mass bins; see Section~\ref{app:metrics}) to verify that improvements are robust to the choice of binning strategy.
Table~\ref{tab:main_results} in the main paper summarizes the same four models with ECE/Brier/AUROC and cost; this table adds AdaECE and supervised baselines.

\begin{table*}[htbp]
\centering
\footnotesize
\begin{tabular}{llccccc}
\toprule
\textbf{Model} & \textbf{Method}
  & \textbf{ECE}$\downarrow$ & \textbf{AdaECE}$\downarrow$ & \textbf{Brier}$\downarrow$
  & \textbf{AUROC}$\uparrow$ & \textbf{Acc} \\
\midrule
\multirow{6}{*}{Llama 3.2-3B}
  & Verbalized     & .170 & .167 & .292 & .510 & .576 \\
  & Self-Consistency$^\dagger$  & .093 & .096 & \textbf{.211} & \textbf{.728} & .624 \\
  & ~~+ Temp Scaling$^*$ & .047 & .075 & .250 & .504 & .576 \\
  & P(True) Norm        & .065 & .067 & .223 & .694 & .576 \\
  & ~~+ Temp Scaling$^*$ & \textbf{.029} & \textbf{.030} & .218 & .692 & .576 \\
  & SECL (Ours)         & \underline{.050} & \underline{.060} & .241 & .587 & .577 \\
\midrule
\multirow{5}{*}{Gemma 2-2B}
  & Verbalized     & .256 & .252 & .314 & .558 & .516 \\
  & ~~+ Temp Scaling$^*$ & .047 & .037 & .249 & .550 & .516 \\
  & P(True) Norm        & .141 & .142 & .259 & \textbf{.650} & .516 \\
  & ~~+ Temp Scaling$^*$ & \textbf{.037} & \textbf{.033} & \textbf{.233} & .647 & .516 \\
  & SECL (Ours)         & \underline{.056} & \underline{.060} & .254 & .548 & .515 \\
\midrule
\multirow{5}{*}{Phi 3.5-Mini (3.8B)}
  & Verbalized     & .251 & .240 & .275 & .600 & .667 \\
  & ~~+ Temp Scaling$^*$ & \textbf{.047} & \textbf{.049} & .215 & .583 & .667 \\
  & P(True) Norm        & .154 & .151 & .227 & \textbf{.675} & .667 \\
  & ~~+ Temp Scaling$^*$ & .059 & .060 & \textbf{.205} & .664 & .667 \\
  & SECL (Ours)         & \underline{.110} & \underline{.119} & .251 & .521 & .665 \\
\midrule
\multirow{5}{*}{Llama 3.1-8B}
  & Verbalized     & .225 & .225 & .258 & .684 & .644 \\
  & ~~+ Temp Scaling$^*$ & \textbf{.080} & .089 & .213 & .681 & .644 \\
  & P(True) Norm        & .120 & .117 & .211 & \textbf{.718} & .646 \\
  & ~~+ Temp Scaling$^*$ & .108 & .103 & \textbf{.208} & .716 & .646 \\
  & SECL (Ours)         & \underline{.083} & \textbf{.080} & .222 & .643 & .646 \\
\bottomrule
\end{tabular}
\vspace{0.3em}
\par\noindent{\footnotesize $^*$Requires ground-truth labels; fitted via 5-fold cross-validation. See Section~\ref{app:posthoc}.}
\par\noindent{\footnotesize $^\dagger$Self-Consistency requires $N{=}10$ sampling
passes per question. We report it for Llama~3.2-3B only to
contextualize SECL's single-pass efficiency; it is omitted for
other models as it measures generation consistency rather than
calibration directly.}
\caption{Overall results across all models and methods (2{,}000 questions each).
\textbf{Bold}: best overall per model for ECE, AdaECE, Brier, and AUROC.
\underline{Underline}: best among label-free methods.}
\label{tab:main_results_full}
\end{table*}

\subsection{Llama Multi-Seed Robustness} \label{app:llama_multiseed}

To quantify seed sensitivity for the main SECL setting, we report Llama~3.2-3B results for seeds 42, 43, and 44 on the same 2{,}000-question sequential stream for both the verbalized baseline and SECL.

\begin{table}[htbp]
\centering
\small
\resizebox{\columnwidth}{!}{%
\begin{tabular}{lcccc}
\toprule
\textbf{Metric} & \textbf{Seed 42} & \textbf{Seed 43} & \textbf{Seed 44} & \textbf{Mean$\pm$Std} \\
\midrule
Verbalized ECE & .1695 & .1757 & .1757 & .1736$\pm$.0029 \\
SECL ECE & .0496 & .0502 & .0392 & \textbf{.0464$\pm$.0051} \\
\midrule
Verbalized AUROC & .5095 & .5119 & .5119 & .5111$\pm$.0011 \\
SECL AUROC & .5870 & .6005 & .5920 & \textbf{.5932$\pm$.0056} \\
\bottomrule
\end{tabular}
}
\caption{Llama~3.2-3B multi-seed robustness (seeds 42, 43, 44). SECL remains strongly better calibrated than the verbalized baseline across seeds and also improves AUROC on this model.}
\label{tab:llama_multiseed}
\end{table}

\section{Post-Hoc Calibration Baselines} \label{app:posthoc}

We compare SECL against standard post-hoc calibration methods:
\textbf{temperature scaling} \citep{guo_calibration_2017} (single parameter $T$
fitted to minimize NLL) and \textbf{Platt scaling} \citep{platt_probabilistic_1999}
(logistic regression $\sigma(a \cdot c + b)$).
Both are applied to the soft verbalized confidence and fitted via 5-fold 
cross-validation (each fold uses 1{,}600 labeled examples for fitting).

\paragraph{Key difference from SECL.}
Temperature and Platt scaling are \emph{supervised} post-hoc methods:
they require ground-truth correctness labels to fit their parameters.
SECL is \emph{unsupervised} at test time; it uses only the model's own 
P(True) signal, with no access to ground truth.
Post-hoc methods are also monotone transforms that 
\emph{cannot} improve discrimination (AUROC);
SECL can improve AUROC by adapting the model's representations.

\begin{table*}[htbp]
\centering
\footnotesize
\begin{tabular}{llccccc}
\toprule
\textbf{Model} & \textbf{Method}
  & \textbf{ECE}$\downarrow$ & \textbf{AdaECE}$\downarrow$
  & \textbf{Brier}$\downarrow$ & \textbf{AUROC}$\uparrow$ 
  & \textbf{Conf.\ range} \\
\midrule
\multirow{4}{*}{Llama 3.2-3B}
  & Verbalized + Temp ($T{=}17.6$)  & .047 & .075 & .250 & .504 & [.44,\,.56] \\
  & Verbalized + Platt              & .021 & .057 & .244 & .481 & [.56,\,.61] \\
  & P(True) Norm + Temp ($T{=}1.8$)  & .029 & .030 & .218 & .692 & [.14,\,.86] \\
  & SECL (no labels)        & .050 & .060 & .241 & .587 & [.05,\,.85] \\
\midrule
\multirow{4}{*}{Gemma 2-2B}
  & Verbalized + Temp ($T{=}11.2$)  & .047 & .037 & .249 & .550 & [.41,\,.59] \\
  & Verbalized + Platt              & .035 & .050 & .250 & .550 & [.42,\,.55] \\
  & P(True) Norm + Temp ($T{=}3.0$)  & .037 & .033 & .233 & .647 & [.26,\,.74] \\
  & SECL (no labels)        & .056 & .060 & .254 & .548 & [.05,\,.95] \\
\midrule
\multirow{4}{*}{Phi 3.5-Mini (3.8B)}
  & Verbalized + Temp ($T{=}3.4$)   & .047 & .049 & .215 & .583 & [.29,\,.71] \\
  & Verbalized + Platt              & .052 & .050 & .216 & .585 & [.16,\,.69] \\
  & P(True) Norm + Temp ($T{=}2.4$)  & .059 & .060 & .205 & .664 & [.21,\,.79] \\
  & SECL (no labels)        & .110 & .119 & .251 & .521 & [.05,\,.95] \\
\midrule
\multirow{4}{*}{Llama 3.1-8B}
  & Verbalized + Temp ($T{=}2.8$)   & .080 & .089 & .213 & .681 & [.35,\,.74] \\
  & Verbalized + Platt              & .063 & .052 & .206 & .677 & [.02,\,.75] \\
  & P(True) Norm + Temp ($T{=}0.7$)  & .108 & .103 & .208 & .716 & [.02,\,.99] \\
  & SECL (no labels)        & .083 & .080 & .222 & .643 & [.25,\,.95] \\
\bottomrule
\end{tabular}
\caption{Post-hoc calibration baselines (5-fold CV). 
Temperature scaling compresses the confidence range substantially,
achieving low ECE by predicting near-constant values for models with
weak verbalized discrimination (fitted $T$ shown).}
\label{tab:posthoc}
\end{table*}

\paragraph{Discussion.}
For Llama, temperature scaling on the soft verbalized confidence requires 
$T{=}17.6$, compressing all predictions to [0.44,\,0.56], essentially 
predicting the base rate.
Platt scaling is more extreme: its output range [0.56,\,0.61] is near-constant,
yielding ECE$=$0.021 but \emph{degrading} AUROC to 0.481 (below chance).
These methods achieve low ECE by destroying discriminative information.
By contrast, SECL achieves ECE$=$0.050 while \emph{improving} AUROC
from 0.510 to 0.587 (+0.077), i.e., genuine calibration rather
than confidence compression.

For Phi, where the verbalized confidence carries more signal 
(baseline AUROC$=$0.600), temperature scaling uses a more moderate 
$T{=}3.4$ and achieves good calibration (ECE$=$0.047).
The strongest supervised baseline across all models is P(True) Norm~+~Temp,
which combines the discriminative P(True) signal with post-hoc recalibration.

\paragraph{Complementarity: SECL + Temperature Scaling.}
SECL and post-hoc calibration are complementary: applying temperature 
scaling to SECL's output further reduces ECE, since SECL improves the 
underlying confidence signal while temperature scaling recalibrates it.

\begin{table}[htbp]
\centering
\small
\resizebox{\columnwidth}{!}{%
\begin{tabular}{llcccc}
\toprule
\textbf{Model} & \textbf{Method}
  & \textbf{ECE}$\downarrow$ & \textbf{AdaECE}$\downarrow$
  & \textbf{Brier}$\downarrow$ & \textbf{AUROC}$\uparrow$ \\
\midrule
\multirow{3}{*}{Llama 3.2-3B}
  & P(True) Norm + Temp$^*$  & .029 & .030 & .218 & .692 \\
  & SECL            & .050 & .060 & .241 & .587 \\
  & SECL + Temp$^*$ & .049 & .053 & .241 & .584 \\
\midrule
\multirow{3}{*}{Gemma 2-2B}
  & P(True) Norm + Temp$^*$  & .037 & .033 & .233 & .647 \\
  & SECL            & .056 & .060 & .254 & .548 \\
  & SECL + Temp$^*$ & .011 & .037 & .249 & .542 \\
\midrule
\multirow{3}{*}{Phi 3.5-Mini (3.8B)}
  & P(True) Norm + Temp$^*$  & .059 & .060 & .205 & .664 \\
  & SECL            & .110 & .119 & .251 & .521 \\
  & SECL + Temp$^*$ & .097 & .082 & .232 & .516 \\
\midrule
\multirow{3}{*}{Llama 3.1-8B}
  & P(True) Norm + Temp$^*$  & .108 & .103 & .208 & .716 \\
  & SECL            & .083 & .080 & .222 & .643 \\
  & SECL + Temp$^*$ & .062 & .045 & .213 & .638 \\
\bottomrule
\end{tabular}}%
\vspace{0.3em}
\newline
{\small $^*$Requires ground-truth labels for temperature fitting.}
\caption{SECL combined with temperature scaling (5-fold CV).
SECL+Temp achieves the best or near-best ECE overall while preserving
SECL's discrimination improvements.}
\label{tab:secl_plus_temp}
\end{table}

\paragraph{Calibration--discrimination trade-off.}
SECL's primary objective is calibration (ECE), not discrimination (AUROC).
Table~\ref{tab:main_results_full} reveals a model-dependent trade-off:
for Llama, SECL \emph{improves} AUROC from .510 to .587 (+.077)
while reducing ECE from .170 to .050,
i.e., the adapted confidence carries better signal.
For Gemma, AUROC is roughly preserved (.558 $\to$ .548, $-$.010)
despite a large ECE reduction (.256 $\to$ .056).
For Phi, the trade-off is more pronounced:
AUROC drops from .600 to .521 ($-$.079) as ECE improves from .251 to .110.
Brier score, which jointly captures calibration and
discrimination, improves for all four models
(Llama~3B: .292 $\to$ .241; Llama~8B: .258 $\to$ .222; Gemma: .314 $\to$ .254; Phi: .275 $\to$ .251),
so the net effect of SECL is positive.
By comparison, supervised temperature scaling also reduces AUROC
(Llama: $-$.006; Gemma: $-$.008; Phi: $-$.017)
and Platt scaling degrades it further for Llama ($-$.029, below chance at .481),
demonstrating that AUROC loss is not unique to SECL but is an inherent
cost of recalibrating overconfident models.

\section{P(True) Normalization and the Generation--Discrimination Gap} \label{app:ptrue_norm}

A key premise of SECL is that LLMs possess a \emph{discriminative} 
confidence signal, P(True), that is better calibrated than their 
\emph{generative} verbalized confidence.
Table~\ref{tab:ptrue_raw_vs_norm} demonstrates this gap and the 
effect of normalization.

\paragraph{Raw P(True) vs.\ the verbalized baseline.}
For Llama, raw P(True) (the model's probability that its own answer 
is correct, without normalization) already improves calibration over 
the verbalized baseline (ECE: .161 vs.\ .170) and dramatically improves 
discrimination (AUROC: .673 vs.\ .510).
This confirms that the model "knows more than it says"; its 
internal judgment of correctness is more reliable than its 
expressed confidence.

\paragraph{Normalization with distractors.}
P(True) Norm applies a temperature-scaled softmax over the correct 
answer and $k{=}4$ distractors, converting a raw probability into a 
calibrated confidence score.
This yields a large further improvement (Llama ECE: .161 $\to$ .065 
at $\tau{=}0.7$), demonstrating that relative comparison against 
plausible alternatives sharpens the discriminative signal.

\begin{table}[htbp]
\centering
\small
\resizebox{\columnwidth}{!}{%
\begin{tabular}{lcccc}
\toprule
\textbf{Method}
  & \textbf{ECE}$\downarrow$ & \textbf{Brier}$\downarrow$
  & \textbf{AUROC}$\uparrow$ & \textbf{Acc} \\
\midrule
Verbalized        & .170 & .292 & .510 & .576 \\
Raw P(True)            & .161 & .248 & .673 & .576 \\
P(True) Norm $\tau{=}0.3$ & .118 & .238 & .685 & .576 \\
P(True) Norm $\tau{=}0.7$ & \textbf{.065} & \textbf{.223} & \textbf{.694} & .576 \\
P(True) Norm $\tau{=}1.5$ & .169 & .254 & .658 & .579 \\
\bottomrule
\end{tabular}}%
\caption{Effect of P(True) normalization (Llama~3.2-3B, $N{=}2{,}000$).
Raw P(True) improves over the verbalized baseline. Normalization with 
temperature $\tau$ yields further gains.
The optimal $\tau$ varies by model (see Table~\ref{tab:tau_sweep}).}
\label{tab:ptrue_raw_vs_norm}
\end{table}

\paragraph{Temperature selection.}
The normalization temperature $\tau$ controls sharpening ($\tau{<}1$) 
or smoothing ($\tau{>}1$) of the softmax over candidates.
Table~\ref{tab:tau_sweep} shows the P(True) Norm ECE at different 
$\tau$ values for each model.
Llama~3.2-3B benefits from sharpening ($\tau{=}0.7$), Gemma and Phi 
require smoothing ($\tau{=}1.5$), and Llama~3.1-8B requires stronger
smoothing ($\tau{=}3.0$), likely reflecting differences in 
the models' internal confidence distributions.
The selected $\tau$ is used in all subsequent experiments.

\begin{table}[htbp]
\centering
\small
\resizebox{\columnwidth}{!}{%
\begin{tabular}{lcccc}
\toprule
$\boldsymbol{\tau}$ & \textbf{Llama 3.2-3B} & \textbf{Gemma 2-2B} & \textbf{Phi 3.5-Mini} & \textbf{Llama 3.1-8B} \\
\midrule
0.3  & .118 & .255 & .172 & .270 \\
0.7  & \textbf{.065} & .215 & .168 & .234 \\
1.0  & .083 & .180 & .161 & .195 \\
1.5  & .169 & \textbf{.141} & \textbf{.154} & .153 \\
3.0  & .247 & .143 & .245 & \textbf{.120} \\
\bottomrule
\end{tabular}}%
\caption{P(True) Norm ECE as a function of normalization 
temperature $\tau$. Bold: selected value per model.}
\label{tab:tau_sweep}
\end{table}

\section{Hyperparameter Sensitivity} \label{app:hp_sensitivity}

Table~\ref{tab:hp_sensitivity} reports ECE and Adaptive ECE on 
Llama~3.2-3B when varying individual hyperparameters.
The $\alpha_{\text{step}}$ and $\delta$ ablations use the full pipeline 
\emph{without} entropy gating (all questions trained) to isolate the loss 
function's effect; the burst size $B$ ablation uses the entropy-gated pipeline.

\begin{table}[htbp]
\centering
\small
\resizebox{\columnwidth}{!}{%
\begin{tabular}{llcc}
\toprule
\textbf{Parameter} & \textbf{Value} & \textbf{ECE}$\downarrow$ & \textbf{AdaECE}$\downarrow$ \\
\midrule
\multirow{3}{*}{$\alpha_{\text{step}}$$^\dagger$}
  & 0.2 & .067 & .083 \\
  & 0.3 & .066 & .069 \\
  & \textbf{0.5} & \textbf{.052} & \textbf{.073} \\
\midrule
\multirow{2}{*}{$\delta$$^\dagger$}
  & \textbf{0.15} & \textbf{.052} & \textbf{.073} \\
  & 0.20 & .064 & .067 \\
\midrule
\multirow{1}{*}{Loss$^\dagger$}
  & Plain MSE (no directional) & .085 & .088 \\
\midrule
\multirow{2}{*}{Burst $B$$^\ddagger$}
  & 20 & .114 & -- \\
  & \textbf{50} & \textbf{.050} & \textbf{.060} \\
\midrule
\multirow{3}{*}{$\tau$ (per model)}
  & Llama: \textbf{0.7} & \textbf{.065} & \textbf{.064} \\
  & Gemma: \textbf{1.5} & .141 & .144 \\
  & Phi: \textbf{1.5} & .154 & .151 \\
\bottomrule
\end{tabular}}%
\caption{Hyperparameter sensitivity (Llama~3.2-3B).
Bold indicates the selected configuration.
$^\dagger$Without entropy gating.
$^\ddagger$With entropy gating.}
\label{tab:hp_sensitivity}
\end{table}

In additional non-default sweeps, we observed further gains for some models
(e.g., Llama in Table~\ref{tab:layer_ablation} and Phi under a tuned
$\lambda$/learning-rate pair, best observed ECE $0.097$). We keep the canonical
defaults in all headline comparisons to preserve a single cross-model protocol
and avoid per-model post-hoc tuning.

\newpage
\onecolumn
\section{Per-Domain Results} \label{app:per_domain}

\begin{table*}[ht]
\centering
\footnotesize
\begin{tabular}{lllccccc}
\toprule
\textbf{Model} & \textbf{Method} & \textbf{Domain} & \textbf{ECE}$\downarrow$ & \textbf{AdaECE}$\downarrow$ & \textbf{Brier}$\downarrow$ & \textbf{AUROC}$\uparrow$ & \textbf{Acc} \\
\midrule
\multirow{12}{*}{Llama 3.2-3B} & \multirow{4}{*}{Verbalized} 
  & GSM8K      & .218 & .215 & .294 & .565 & .472 \\
& & MMLU       & .106 & .109 & .253 & .601 & .562 \\
& & ARC        & \textbf{.095} & .112 & .207 & .568 & .726 \\
& & TruthfulQA & .372 & .371 & .414 & .301 & .544 \\
\cmidrule{2-8}
& \multirow{4}{*}{P(True) Norm} 
  & GSM8K      & .133 & .138 & .264 & .594 & .476 \\
& & MMLU       & .091 & .090 & .239 & .652 & .566 \\
& & ARC        & .117 & .132 & .210 & .624 & .728 \\
& & TruthfulQA & .089 & \textbf{.092} & .180 & .818 & .532 \\
\cmidrule{2-8}
& \multirow{4}{*}{SECL} 
  & GSM8K      & \textbf{.070} & \textbf{.079} & .249 & .573 & .488 \\
& & MMLU       & \textbf{.067} & \textbf{.060} & .249 & .549 & .558 \\
& & ARC        & .112 & \textbf{.106} & .207 & .616 & .714 \\
& & TruthfulQA & \textbf{.068} & .114 & .260 & .454 & .546 \\
\midrule
\multirow{12}{*}{Gemma 2-2B} & \multirow{4}{*}{Verbalized} 
  & GSM8K      & .549 & .549 & .446 & .658 & .186 \\
& & MMLU       & .194 & .178 & .271 & .608 & .576 \\
& & ARC        & \textbf{.082} & \textbf{.085} & .201 & .534 & .738 \\
& & TruthfulQA & .267 & .263 & .338 & .417 & .566 \\
\cmidrule{2-8}
& \multirow{4}{*}{P(True) Norm} 
  & GSM8K      & .395 & .394 & .326 & .685 & .186 \\
& & MMLU       & .163 & .175 & .263 & .616 & .576 \\
& & ARC        & .185 & .182 & .229 & .597 & .738 \\
& & TruthfulQA & \textbf{.104} & \textbf{.104} & .213 & .729 & .566 \\
\cmidrule{2-8}
& \multirow{4}{*}{SECL} 
  & GSM8K      & \textbf{.356} & \textbf{.356} & .271 & .609 & .180 \\
& & MMLU       & \textbf{.054} & \textbf{.064} & .238 & .594 & .578 \\
& & ARC        & .144 & .136 & .213 & .591 & .728 \\
& & TruthfulQA & .210 & .198 & .293 & .343 & .574 \\
\midrule
\multirow{12}{*}{Phi 3.5-Mini} & \multirow{4}{*}{Verbalized} 
  & GSM8K      & .290 & .290 & .304 & .563 & .650 \\
& & MMLU       & .264 & .257 & .286 & .602 & .648 \\
& & ARC        & \textbf{.109} & \textbf{.105} & .150 & .588 & .826 \\
& & TruthfulQA & .343 & .336 & .362 & .592 & .546 \\
\cmidrule{2-8}
& \multirow{4}{*}{P(True) Norm} 
  & GSM8K      & \textbf{.261} & \textbf{.257} & .303 & .559 & .650 \\
& & MMLU       & .171 & .158 & .243 & .648 & .648 \\
& & ARC        & .129 & .119 & .157 & .680 & .826 \\
& & TruthfulQA & \textbf{.146} & \textbf{.141} & .205 & .771 & .546 \\
\cmidrule{2-8}
& \multirow{4}{*}{SECL} 
  & GSM8K      & .283 & .280 & .297 & .575 & .658 \\
& & MMLU       & \textbf{.054} & \textbf{.058} & .223 & .604 & .644 \\
& & ARC        & .129 & .129 & .167 & .499 & .826 \\
& & TruthfulQA & .229 & .224 & .319 & .407 & .532 \\
\bottomrule
\end{tabular}
\caption{Per-domain results for Llama 3.2-3B, Gemma 2-2B, and Phi 3.5-Mini. Best ECE and AdaECE per domain are in \textbf{bold}.}
\label{tab:per_domain_combined}
\end{table*}

\clearpage

\twocolumn
\section{Ablation Studies} \label{app:ablations}

\paragraph{Weight accumulation.}
By default, SECL accumulates LoRA weights across the stream
(each question's adaptation builds on the previous state).
Table~\ref{tab:ablation_accum} shows that disabling accumulation
(resetting LoRA to the base model after each question)
is catastrophic: ECE degrades from 0.050 to 0.237, and AUROC
drops below chance (0.484), i.e., isolated per-question updates destroy the model's confidence
signal.

\begin{table}[htbp]
\centering
\small
\resizebox{\columnwidth}{!}{%
\begin{tabular}{lcccc}
\toprule
\textbf{Setting}
  & \textbf{ECE}$\downarrow$ & \textbf{Brier}$\downarrow$
  & \textbf{AUROC}$\uparrow$ & \textbf{Acc} \\
\midrule
SECL (accumulate)  & \textbf{.050} & \textbf{.241} & \textbf{.587} & .577 \\
Reset per question & .237 & .327 & .484 & .576 \\
\bottomrule
\end{tabular}}%
\caption{Weight accumulation ablation (Llama~3.2-3B).}
\label{tab:ablation_accum}
\end{table}

\paragraph{Entropy gating vs.\ always-train.}
Training on every question (no entropy gating) achieves comparable ECE
to SECL (0.047 vs.\ 0.050), but at 4$\times$ the computational cost
since all 2{,}000 questions receive TTT updates.
Entropy gating triggers calibration bursts when distribution shift is detected;
under our protocol this results in 25.6\% of questions receiving TTT updates,
achieving the same calibration with 75\% fewer gradient steps.

\begin{table}[htbp]
\centering
\small
\resizebox{\columnwidth}{!}{%
\begin{tabular}{lcccc}
\toprule
\textbf{Setting}
  & \textbf{ECE}$\downarrow$ & \textbf{Brier}$\downarrow$
  & \textbf{AUROC}$\uparrow$ & \textbf{Trained \%} \\
\midrule
Always-train   & \textbf{.047} & \textbf{.240} & \textbf{.593} & 100\% \\
Entropy-gated (SECL) & .050 & .241 & .587 & 25.6\% \\
\bottomrule
\end{tabular}}%
\caption{Entropy gating vs.\ always-train (Llama~3.2-3B).
Entropy gating matches always-train ECE at a fraction of the cost.}
\label{tab:ablation_phgate}
\end{table}

\paragraph{KL divergence regularization.}
We experiment with adding a KL divergence term
$\beta \cdot D_{\text{KL}}(p_{\text{base}} \| p_{\text{adapted}})$
to the training loss to preserve the base model's output distribution.
Table~\ref{tab:ablation_kl} shows that a small $\beta{=}0.01$ slightly
improves Llama ECE (0.044 vs.\ 0.050) but degrades Gemma and Phi,
while $\beta{=}0.1$ is too aggressive for all models.
We use $\beta{=}0$ (no KL term) in all reported experiments.

\begin{table}[htbp]
\centering
\small
\resizebox{\columnwidth}{!}{%
\begin{tabular}{llcccc}
\toprule
\textbf{Model} & $\boldsymbol{\beta}$
  & \textbf{ECE}$\downarrow$ & \textbf{Brier}$\downarrow$
  & \textbf{AUROC}$\uparrow$ & \textbf{Acc} \\
\midrule
\multirow{3}{*}{Llama 3.2-3B}
  & \textbf{0} (default) & .050 & \textbf{.241} & .587 & \textbf{.577} \\
  & 0.01     & \textbf{.044} & .242 & \textbf{.591} & .573 \\
  & 0.1      & .149 & .279 & .527 & .569 \\
\midrule
\multirow{3}{*}{Gemma 2-2B}
  & \textbf{0} (default) & \textbf{.056} & \textbf{.254} & .548 & .515 \\
  & 0.01     & .127 & .271 & .486 & .514 \\
  & 0.1      & .238 & .305 & \textbf{.551} & \textbf{.520} \\
\midrule
\multirow{3}{*}{Phi 3.5-Mini (3.8B)}
  & \textbf{0} (default) & \textbf{.110} & \textbf{.251} & .521 & .665 \\
  & 0.01     & .144 & .252 & .506 & .669 \\
  & 0.1      & .248 & .272 & \textbf{.598} & \textbf{.673} \\
\bottomrule
\end{tabular}}%
\caption{KL regularization ablation across models (SECL + entropy gating).}
\label{tab:ablation_kl}
\end{table}

\section{Ablation: LoRA Layer Position} \label{app:layer_ablation}

To justify placing LoRA adapters on late transformer layers,
we compare five layer configurations on Llama~3.2-3B while holding
all other hyperparameters fixed (top block of Table~\ref{tab:layer_ablation}).
Mid and late layers achieve the same ECE (0.039), while early layers
remain competitive (0.046).
Adapting all 28 layers yields the best AUROC and Brier
but the worst ECE (0.058), suggesting that adapting too many layers
introduces noise that hurts bin-level calibration despite improving
per-question discrimination.
The robustness across layer positions
indicates that SECL's calibration mechanism is not tightly coupled
to a specific layer group; we use late layers as the default following
evidence that calibration-relevant representations concentrate in
intermediate-to-late layers \citep{du_haloscope_2024}.

We further verify the effect of layer \emph{count} on Gemma~2-2B and
Phi~3.5-Mini (3.8B) (bottom block of Table~\ref{tab:layer_ablation}).
Halving the adapter count from 8 to 4 layers roughly doubles ECE on
both models (Gemma: 0.056$\to$0.116; Phi: 0.110$\to$0.222) with
negligible accuracy change, so 8 late layers provide
sufficient capacity for calibration across architectures.

\begin{table}[htbp]
\centering
\small
\resizebox{\columnwidth}{!}{%
\begin{tabular}{llccccc}
\toprule
\textbf{Model} & \textbf{Layers} & \textbf{Range}
  & \textbf{ECE}$\downarrow$ & \textbf{Brier}$\downarrow$
  & \textbf{AUROC}$\uparrow$ & \textbf{Acc} \\
\midrule
\multirow{5}{*}{Llama 3.2-3B}
  & Early     & 0--3   & .046 & .235 & .619 & .591 \\
  & Mid       & 12--15 & \textbf{.039} & .233 & .628 & .573 \\
  & Late (default) & 24--27 & \textbf{.039} & .241 & .592 & .570 \\
  & Last 8    & 20--27 & .044 & .240 & .609 & .558 \\
  & All       & 0--27  & .058 & \textbf{.232} & \textbf{.640} & \textbf{.588} \\
\midrule
\multirow{2}{*}{Gemma 2-2B}
  & Last 4    & 22--25 & .116 & .277 & .464 & .513 \\
  & Last 8 (default) & 18--25 & \textbf{.056} & \textbf{.254} & \textbf{.548} & \textbf{.515} \\
\midrule
\multirow{2}{*}{Phi 3.5-Mini (3.8B)}
  & Last 4    & 28--31 & .222 & .277 & \textbf{.524} & \textbf{.669} \\
  & Last 8 (default) & 24--31 & \textbf{.110} & \textbf{.251} & .521 & .665 \\
\bottomrule
\end{tabular}}%
\caption{LoRA layer ablation.
\textbf{Top:} layer position on Llama~3.2-3B (4 layers each except "All").
\textbf{Bottom:} layer count on Gemma~2-2B and Phi~3.5-Mini (3.8B).}
\label{tab:layer_ablation}
\end{table}

\section{Reliability Diagrams and Normalization Effect} \label{app:reliability}

\begin{figure*}[htbp]
    \centering
    \includegraphics[width=0.85\textwidth]{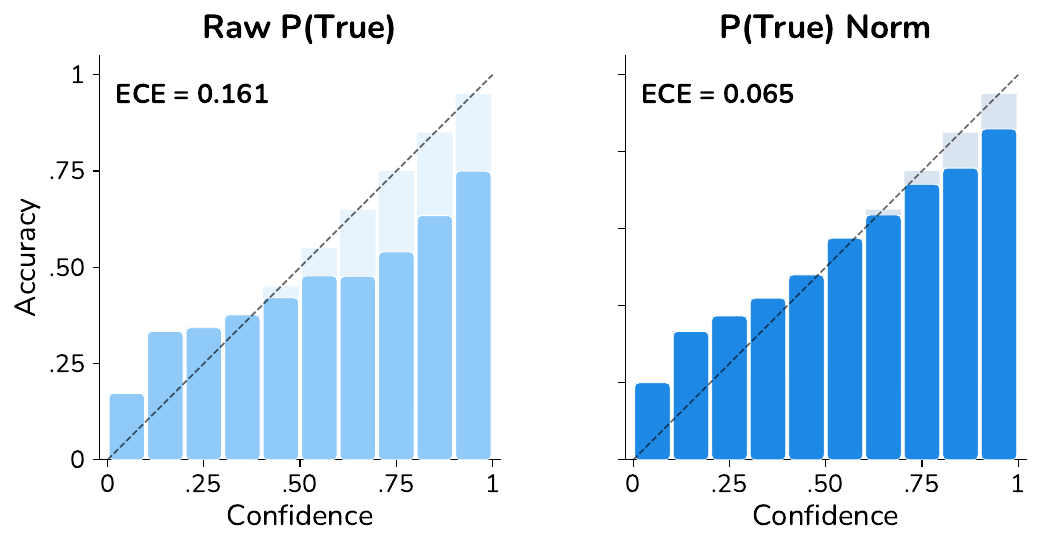}
    \caption{Raw $P(\text{True})$ (\textbf{left}) vs.\
    distractor-normalized $\text{NormP}_{\text{True}}$
    (\textbf{right}) on Llama~3.2-3B.
    Normalization suppresses suggestibility bias, producing a
    better-calibrated supervision signal
    (ECE: 0.161\,$\to$\,0.065).}
    \label{fig:ptrue_norm}
\end{figure*}

Figure~\ref{fig:ptrue_norm} compares reliability diagrams for raw
$P(\text{True})$ and distractor-normalized
$\text{NormP}_{\text{True}}$ on Llama~3.2-3B.
Raw $P(\text{True})$ exhibits upward bias in mid-to-high confidence
bins, consistent with the suggestibility effect documented by
\citet{wang_calibrating_2025}: the model tends to affirm
plausible-sounding answers regardless of correctness.
Distractor normalization removes this bias by converting
absolute affirmation into relative preference among candidate
answers, reducing ECE from 0.161 to 0.065.

\begin{figure*}[htbp]
    \centering
    \includegraphics[width=0.85\textwidth]{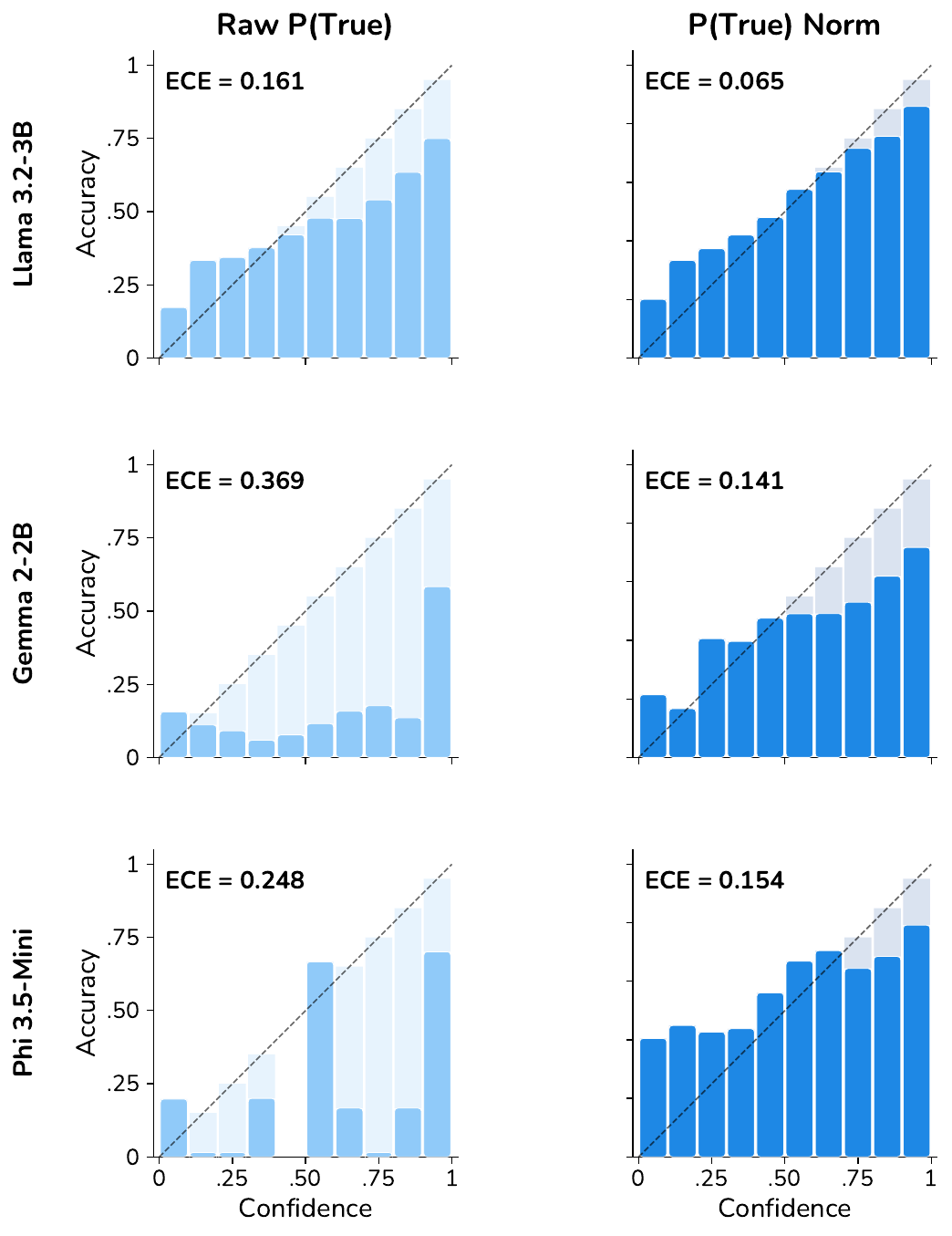}
    \caption{Raw $P(\text{True})$ (\textbf{left}) vs.\ $\text{NormP}_{\text{True}}$ (\textbf{right}) for the three non-8B models. Distractor normalization consistently reduces calibration error across model families.}
    \label{fig:ptrue_norm_all}
\end{figure*}

Figure~\ref{fig:ptrue_norm_all} extends this comparison to all three
models. Gemma~2-2B benefits the most (ECE: 0.369\,$\to$\,0.141),
while Phi~3.5-Mini (3.8B) shows a similarly large reduction
(ECE: 0.248\,$\to$\,0.154).

\begin{figure*}[htbp]
    \centering
    \includegraphics[width=0.85\textwidth]{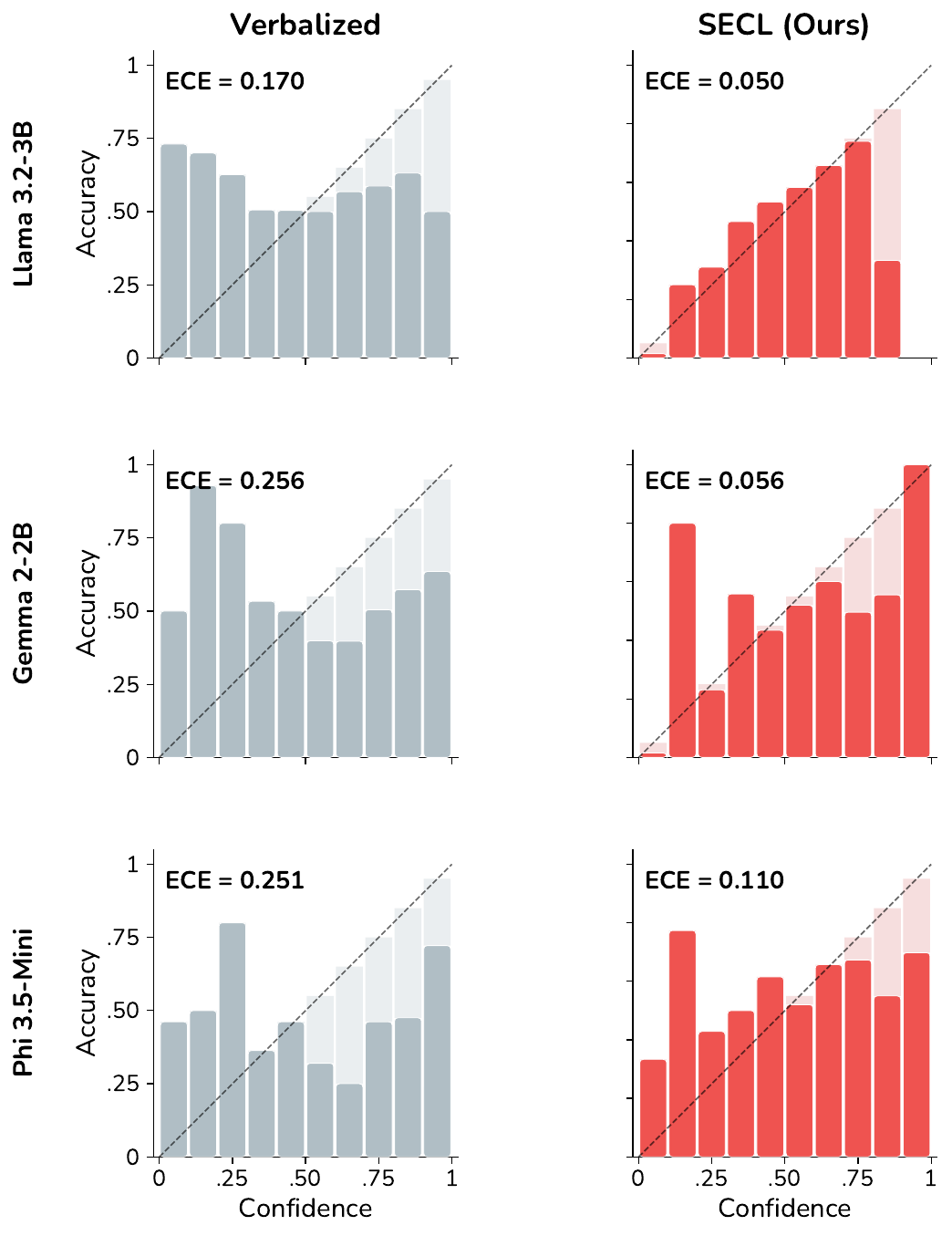}
    \caption{Reliability diagrams: Verbalized baseline (\textbf{left}) vs.\ SECL (\textbf{right}) for Llama~3.2-3B (ECE: 0.170\,$\to$\,0.050), Gemma~2-2B (ECE: 0.256\,$\to$\,0.056), and Phi~3.5-Mini (3.8B) (ECE: 0.251\,$\to$\,0.110). SECL consistently shifts predictions toward the diagonal, eliminating the systematic overconfidence visible in the verbalized baselines.}
    \label{fig:rel_all}
\end{figure*}

Reliability diagrams for the verbalized baseline vs.\ SECL on all three
models are shown in Figure~\ref{fig:rel_all}.

\section{Domain Order Sensitivity} \label{app:domain_order}

To assess sensitivity to the presentation order of domains, we run SECL 
with the reversed sequence 
(TruthfulQA $\to$ ARC $\to$ MMLU $\to$ GSM8K).
Table~\ref{tab:domain_order_full} compares overall metrics; 
Table~\ref{tab:domain_order_perdomain} gives per-domain detail.
SECL improves over the verbalized baseline under both orderings for all models,
though the original order yields lower ECE.
The gap is smallest for Llama~3.2-3B (0.050 vs.\ 0.068); Llama~3.1-8B shows
a larger gap (0.083 vs.\ 0.189), partly because the reversed order triggers
gating on 62.6\% of questions (vs.\ 12.6\% forward), amplifying cumulative
LoRA drift on the longer model.

\begin{table}[htbp]
\centering
\small
\resizebox{\columnwidth}{!}{%
\begin{tabular}{llccccc}
\toprule
\textbf{Model} & \textbf{Order}
  & \textbf{ECE}$\downarrow$ & \textbf{AdaECE}$\downarrow$ & \textbf{Brier}$\downarrow$
  & \textbf{AUROC}$\uparrow$ & \textbf{Acc} \\
\midrule
\multirow{3}{*}{Llama 3.2-3B}
  & Verbalized       & .170 & .167 & .292 & .510 & .576 \\
  & SECL $\mathcal{O}$ & \textbf{.050} & \textbf{.060} & .241 & .587 & .577 \\
  & SECL $\mathcal{R}$ & .068 & .082 & .248 & .578 & .575 \\
\midrule
\multirow{3}{*}{Gemma 2-2B}
  & Verbalized       & .256 & .252 & .314 & .558 & .516 \\
  & SECL $\mathcal{O}$ & \textbf{.056} & \textbf{.060} & .254 & .548 & .515 \\
  & SECL $\mathcal{R}$ & .149 & .133 & .258 & .625 & .520 \\
\midrule
\multirow{3}{*}{Phi 3.5-Mini (3.8B)}
  & Verbalized       & .251 & .240 & .275 & .600 & .667 \\
  & SECL $\mathcal{O}$ & \textbf{.110} & \textbf{.119} & .251 & .521 & .665 \\
  & SECL $\mathcal{R}$ & .173 & .170 & .250 & .575 & .674 \\
\midrule
\multirow{3}{*}{Llama 3.1-8B}
  & Verbalized       & .225 & .225 & .258 & .684 & .644 \\
  & SECL $\mathcal{O}$ & \textbf{.083} & \textbf{.080} & .222 & .643 & .646 \\
  & SECL $\mathcal{R}$ & .189 & .191 & .272 & .580 & .629 \\
\bottomrule
\end{tabular}}%
\caption{Domain order sensitivity (overall metrics).
$\mathcal{O}$: GSM8K$\to$MMLU$\to$ARC$\to$TruthfulQA.
$\mathcal{R}$: TruthfulQA$\to$ARC$\to$MMLU$\to$GSM8K.}
\label{tab:domain_order_full}
\end{table}

\begin{table}[htbp]
\centering
\small
\resizebox{\columnwidth}{!}{%
\begin{tabular}{llcccc}
\toprule
\textbf{Model} & \textbf{Order}
  & \textbf{GSM8K} & \textbf{MMLU}
  & \textbf{ARC} & \textbf{TruthfulQA} \\
\midrule
\multirow{2}{*}{Llama 3.2-3B}
  & $\mathcal{O}$ & .070 & .067 & .112 & .068 \\
  & $\mathcal{R}$ & .117 & .032 & .081 & .134 \\
\midrule
\multirow{2}{*}{Gemma 2-2B}
  & $\mathcal{O}$ & .356 & .054 & .144 & .210 \\
  & $\mathcal{R}$ & .396 & .096 & .046 & .189 \\
\midrule
\multirow{2}{*}{Phi 3.5-Mini (3.8B)}
  & $\mathcal{O}$ & .283 & .054 & .129 & .229 \\
  & $\mathcal{R}$ & .068 & .213 & .099 & .343 \\
\midrule
\multirow{2}{*}{Llama 3.1-8B}
  & $\mathcal{O}$ & .066 & .120 & .198 & .038 \\
  & $\mathcal{R}$ & .206 & .217 & .234 & .223 \\
\bottomrule
\end{tabular}}%
\caption{Per-domain ECE for original ($\mathcal{O}$) vs.\ reversed 
($\mathcal{R}$) domain order.}
\label{tab:domain_order_perdomain}
\end{table}

\paragraph{ARC domain ordering effect.}
ARC-Challenge is the only domain where SECL consistently increases ECE
relative to the verbalized baseline in the forward order
(Llama~3B: $.095 \to .112$; Gemma: $.082 \to .144$; Phi: $.109 \to .129$; Llama~8B: $.108 \to .198$).
In the reversed order, ARC ECE \emph{improves} for the three smaller models
(Llama~3B: $.095 \to .081$; Gemma: $.082 \to .046$; Phi: $.109 \to .099$),
though Llama~8B degrades further ($.108 \to .234$).
For the smaller models, the ARC degradation is therefore an artifact of domain sequencing, specifically,
the LoRA weights adapted on preceding domains transfer poorly to ARC's
scientific reasoning format, not a limitation of SECL itself.
Overall ECE remains improved under both orderings for all models,
so the per-domain redistribution does not undermine the method's aggregate benefit.

\section{Computational Cost Analysis} \label{app:cost}

Table~\ref{tab:cost} compares the computational cost of SECL against
running P(True) Norm on every question.
We report costs in forward-pass equivalents (FWD-eq), where one 
backward pass $\approx$ 2 FWD.
For questions where entropy gating skips TTT, SECL requires only 1~FWD 
(the generation pass); the P(True) Norm baseline requires $1 + (k{+}1) = 6$~FWD 
per question ($k{=}4$ distractors plus the correct answer).

\begin{table}[htbp]
\centering
\small
\resizebox{\columnwidth}{!}{%
\begin{tabular}{lcccc}
\toprule
\textbf{Model} & \textbf{Trained} & \textbf{Skipped}
  & \textbf{SECL} & \textbf{P(True)} \\
\midrule
Llama 3.2-3B & 512 (25.6\%) & 1{,}488 & 9{,}168 & 12{,}000 \\
Llama 3.1-8B & 251 (12.6\%) & 1{,}749 & 5{,}514 & 12{,}000 \\
Gemma 2-2B   & 119 ~~(5.9\%) & 1{,}881 & 3{,}666 & 12{,}000 \\
Phi 3.5-Mini (3.8B) & 160 ~~(8.0\%) & 1{,}840 & 4{,}240 & 12{,}000 \\
\bottomrule
\end{tabular}}%
\caption{Computational cost comparison (in FWD-equivalents).
\emph{Trained} = questions receiving TTT;
\emph{Skipped} = entropy-gated questions (generation only).
Per trained question: ${\approx}15$ FWD-eq 
(1~gen $+$ 5~P(True) $+$ 3~epochs $\times$ 3~FWD-eq/step).
Per skipped question: $1$ FWD-eq.
P(True) Norm baseline: 6 FWD-eq $\times$ 2{,}000 $= 12{,}000$.}
\label{tab:cost}
\vspace{-0.5em}
\end{table}

SECL is cheaper than P(True) Norm baseline for all four models.
Gemma and Phi trigger infrequently (6--8\%), reducing cost to roughly
a third of the baseline while achieving substantially better calibration.
Even Llama, which triggers most often (25.6\%), remains below the
baseline cost while improving ECE from .065 to .050.
The temperature scaling and Platt scaling add negligible compute
(a single parameter fit) but require ground-truth labels, which are 
unavailable in the test-time setting.

\section{Negative Control: Qwen 2.5-3B} \label{app:qwen}

We evaluate Qwen~2.5-3B-Instruct \citep{qwen_qwen25_2025} as a negative control.
Unlike the other four models, Qwen's P(True) Norm baseline 
(\emph{best} $\tau{=}1.0$, ECE$=$0.257) is \emph{worse} than its
verbalized baseline (ECE$=$0.247), indicating that the 
discriminative signal provides no calibration advantage over generation.
Since SECL distills the discriminative signal into generative confidence,
the method cannot improve calibration when the discriminative signal 
itself is uninformative.
This confirms a key prerequisite of our approach: SECL succeeds 
precisely when a generation--discrimination gap exists.

\begin{table}[htbp]
\centering
\small
\begin{tabular}{lc}
\toprule
\textbf{Method} & \textbf{ECE}$\downarrow$ \\
\midrule
Verbalized & \textbf{.247} \\
\addlinespace
P(True) Norm $\tau{=}0.3$ & .290 \\
P(True) Norm $\tau{=}0.7$ & .263 \\
P(True) Norm $\tau{=}1.0$ & .257 \\
P(True) Norm $\tau{=}1.5$ & .265 \\
P(True) Norm $\tau{=}2.0$ & .267 \\
P(True) Norm $\tau{=}3.0$ & .291 \\
\bottomrule
\end{tabular}
\caption{Qwen~2.5-3B: no generation--discrimination gap.
All P(True) Norm temperatures yield worse ECE than the verbalized baseline.}
\label{tab:qwen}
\end{table}

\clearpage

\section{Entropy Gating Trigger Statistics} \label{app:ph_stats}

\begin{table}[htbp]
\centering
\small
\resizebox{\columnwidth}{!}{%
\begin{tabular}{lccc}
\toprule
\textbf{Model} & \textbf{Total} & \textbf{Trained (\%)} & \textbf{Skipped (\%)} \\
\midrule
Llama 3.2-3B & 2{,}000 & 512 (25.6\%) & 1{,}488 (74.4\%) \\
Gemma 2-2B   & 2{,}000 & 119 ~~(5.9\%) & 1{,}881 (94.1\%) \\
Phi 3.5-Mini (3.8B) & 2{,}000 & 160 ~~(8.0\%) & 1{,}840 (92.0\%) \\
Llama 3.1-8B & 2{,}000 & 251 (12.6\%) & 1{,}749 (87.4\%) \\
\bottomrule
\end{tabular}}%
\caption{Entropy gating trigger statistics per model.
\emph{Trained\%} is the fraction of questions receiving TTT adaptation.}
\label{tab:ph_triggers}
\end{table}

\section{Scaling to Llama 3.1-8B} \label{app:scaling_8b}

To test whether SECL scales to larger models, we evaluate
Llama~3.1-8B-Instruct (8.03B parameters) on the same 2{,}000-question stream.
We use LoRA on the last 8 layers (rank~8), $\tau{=}3.0$,
and the same entropy gating parameters as the 3B models.

\begin{table}[htbp]
\centering
\small
\resizebox{\columnwidth}{!}{%
\begin{tabular}{lccccc}
\toprule
\textbf{Method}
  & \textbf{ECE}$\downarrow$ & \textbf{AdaECE}$\downarrow$ & \textbf{Brier}$\downarrow$
  & \textbf{AUROC}$\uparrow$ & \textbf{Acc} \\
\midrule
Verbalized      & .225 & .225 & .258 & .684 & .644 \\
~~+ Temp Scaling$^*$ & \textbf{.080} & .089 & .213 & .681 & .644 \\
P(True) Norm         & .120 & .117 & .211 & \textbf{.718} & .646 \\
~~+ Temp Scaling$^*$ & .108 & .103 & \textbf{.208} & .716 & .646 \\
SECL (Ours)          & \underline{.083} & \textbf{.080} & .222 & .643 & .646 \\
\bottomrule
\end{tabular}}%
\caption{Llama~3.1-8B results. SECL reduces ECE by 63\% relative
to the verbalized baseline while adapting only 12.6\% of questions.
$^*$Requires ground-truth labels (5-fold CV).}
\label{tab:8b_results}
\end{table}

\begin{table}[htbp]
\centering
\small
\resizebox{\columnwidth}{!}{%
\begin{tabular}{llcccc}
\toprule
\textbf{Method} & \textbf{Domain}
  & \textbf{ECE}$\downarrow$ & \textbf{Brier}$\downarrow$
  & \textbf{AUROC}$\uparrow$ & \textbf{Acc} \\
\midrule
\multirow{4}{*}{Verbalized}
  & GSM8K      & .298 & .305 & .714 & .602 \\
  & MMLU       & .208 & .251 & .644 & .670 \\
  & ARC        & .108 & .164 & .633 & .798 \\
  & TruthfulQA & .287 & .310 & .669 & .506 \\
\midrule
\multirow{4}{*}{SECL}
  & GSM8K      & \textbf{.066} & .238 & .602 & .572 \\
  & MMLU       & \textbf{.120} & .223 & .591 & .686 \\
  & ARC        & .198 & .178 & .632 & .824 \\
  & TruthfulQA & \textbf{.038} & .249 & .552 & .504 \\
\bottomrule
\end{tabular}}%
\caption{Per-domain results for Llama~3.1-8B.}
\label{tab:8b_per_domain}
\end{table}

SECL achieves ECE$=$0.083 on the 8B model, a 63\% reduction from the
verbalized baseline (0.225), confirming that the method scales
effectively. The pattern mirrors the 3B results: large improvements
on GSM8K, MMLU, and TruthfulQA, with ARC remaining resistant to
improvement (see domain ordering analysis in Section~\ref{app:domain_order}).
The entropy gate triggers on 12.6\% of questions, intermediate between
the 3B models' 5.9--25.6\% range.

\section{Ablation: Training Target Signal}
\label{app:target_ablation}

A natural question is whether P(True) Norm is the right pseudo-label
for SECL training.
Self-Consistency \citep[SC;][]{wang_selfconsistency_2023} is an alternative unsupervised signal that
measures agreement among $N{=}10$ temperature-sampled answers.
We run the full SECL pipeline (LoRA adaptation, entropy gating,
burst training, MSE loss) with SC as the training target, keeping
all other hyperparameters identical to the canonical configuration.

\begin{table}[htbp]
\centering
\small
\resizebox{\columnwidth}{!}{%
\begin{tabular}{lcccc}
\toprule
\textbf{Training Target}
  & \textbf{ECE}$\downarrow$ & \textbf{Brier}$\downarrow$
  & \textbf{AUROC}$\uparrow$ & \textbf{Acc} \\
\midrule
None (Verbalized baseline)     & .170 & .292 & .510 & .576 \\
Self-Consistency ($N{=}10$) & .432 & .470 & .443 & .564 \\
P(True) Norm (ours)        & \textbf{.050} & \textbf{.241} & \textbf{.587} & \textbf{.577} \\
\bottomrule
\end{tabular}}%
\caption{Effect of training target on Llama~3.2-3B.
All rows use the same SECL pipeline; only the pseudo-label
differs. Self-Consistency as a target degrades calibration below
the untrained baseline.}
\label{tab:target_ablation}
\end{table}

\paragraph{Why the target signal determines calibration quality.}
SECL minimizes $\mathcal{L} = \mathbb{E}[(c(q) - t(q))^2]$, where
$c(q)$ is the model's verbalized confidence and $t(q)$ is the
training target.
With sufficient model capacity and training convergence,
$c^*(q) \to t(q)$, so the trained model \emph{inherits the
calibration of its target}:
$\mathrm{ECE}(c^*) \to \mathrm{ECE}(t)$.
The choice of target therefore bounds the achievable calibration.

\paragraph{SC is a biased proxy for correctness.}
As $N \to \infty$, SC$(q) \to p_{\mathrm{gen}}(\text{mode} \mid q)$,
the probability mass the generation distribution places on its modal
answer.
For modern LLMs, the generation distribution is low-entropy:
$p_{\mathrm{gen}}(\text{mode} \mid q)$ is typically high
($>$0.7) regardless of whether the answer is correct.
Among questions where SC~$\approx 1$, the actual fraction correct
can be significantly below~1, violating the calibration condition
$P(Y{=}1 \mid c{=}p) = p$.
Training on SC therefore teaches the model to report generation
concentration as confidence, producing systematic overconfidence.

\begin{figure*}[ht]
\centering
\includegraphics[width=0.85\textwidth]{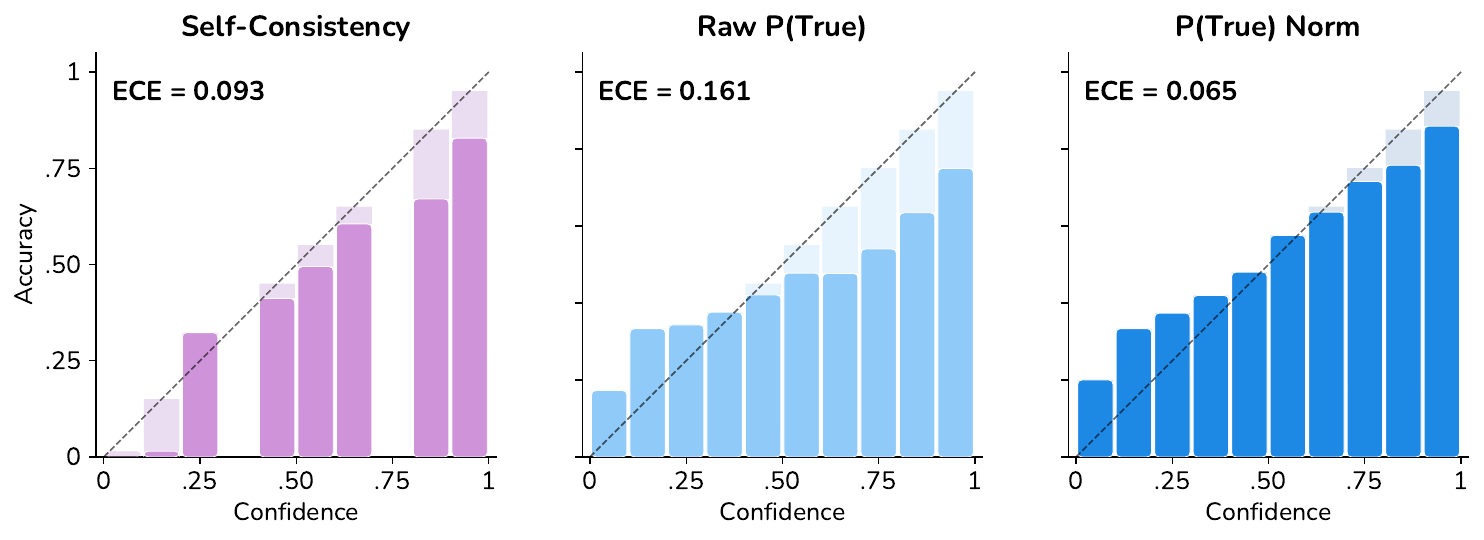}
\caption{Reliability diagrams comparing candidate training targets on Llama~3.2-3B. Self-Consistency (\textbf{left}, ECE$=$0.093) produces a reasonable calibration curve but with systematic bias. Raw P(True) (\textbf{center}, ECE$=$0.161) exhibits upward bias from suggestibility. P(True) Norm (\textbf{right}, ECE$=$0.065) tracks the diagonal most closely, confirming it as the best-calibrated supervision signal for SECL training.}
\label{fig:sc_ptrue_norm_llama}
\end{figure*}

\begin{figure*}[ht]
\centering
\includegraphics[width=0.85\textwidth]{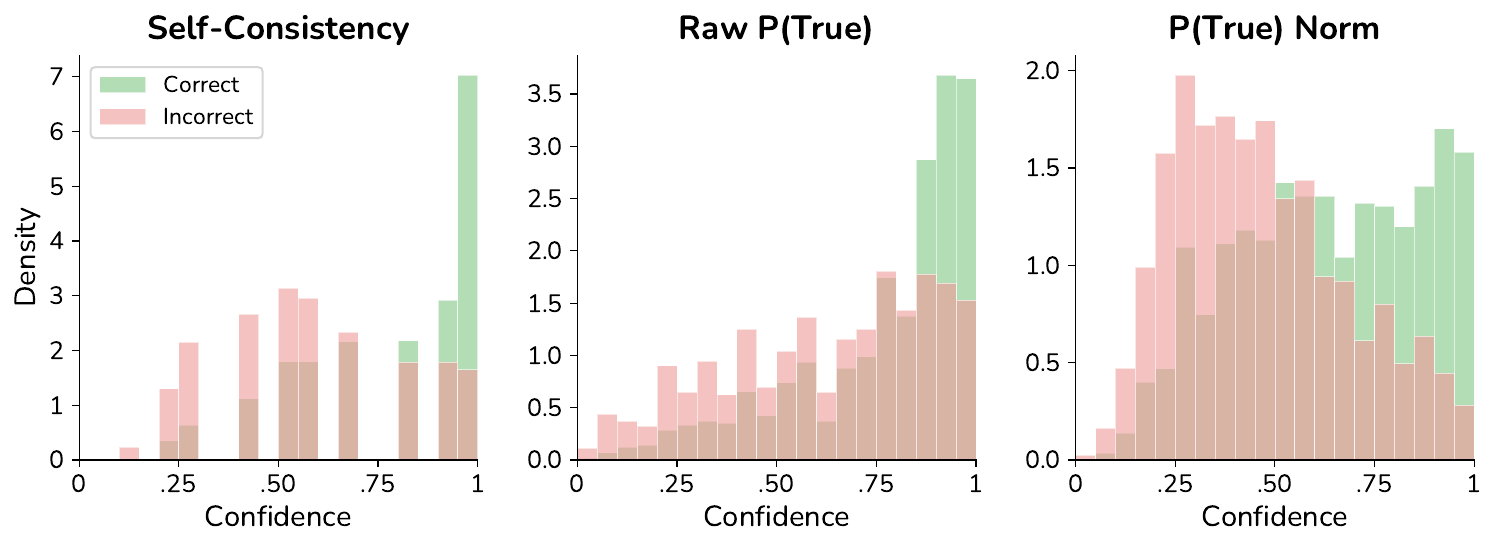}
\caption{Confidence score distributions for correct (green) and incorrect (red) predictions on Llama~3.2-3B. Self-Consistency and Raw P(True) both cluster near 1.0 with poor class separation; P(True) Norm spreads scores across $[0,1]$ with better separation between correct and incorrect predictions.}
\label{fig:target_distributions}
\end{figure*}

\paragraph{P(True) Norm is a less biased proxy.}
P(True) Norm computes $\mathrm{softmax}(\log P(\text{True} \mid
q, a_i) / \tau)_0$ over the model's answer and $k$ distractors.
When the model cannot distinguish its answer from distractors
(equal $P(\text{True})$ scores), P(True) Norm~$\to 1/(k{+}1)$,
correctly indicating low confidence.
When the model strongly prefers its answer,
P(True) Norm~$\to 1$.
This normalization maps the discriminative signal to $[0,1]$, aligning it more closely with the correctness probability and reducing both raw and trained ECE.

The raw P(True) Norm signal (ECE$=$0.065) is better calibrated than the
raw Self-Consistency signal (ECE$=$0.093) \emph{before any training}.
After SECL training with the P(True) Norm target, the model's verbalized
confidence inherits this calibration (ECE$=$0.050), while training
With the SC target, SC's biases are amplified (ECE$=$0.432).
Figure~\ref{fig:sc_ptrue_norm_llama} provides direct visual
evidence via reliability diagrams: SC produces a monotonically
increasing but biased calibration curve, Raw P(True) suffers from
suggestibility bias in mid-to-high bins, while P(True) Norm
tracks the diagonal most closely.
Figure~\ref{fig:target_distributions} complements this with
score distributions, showing that P(True) Norm achieves the best
separation between correct and incorrect predictions.

Calibration quality is thus determined by the training target.
P(True) Norm is a better proxy for $P(\text{correct})$ than Self-Consistency; the adaptation mechanism distills whatever signal it receives, so signal quality determines the ceiling of SECL's effectiveness.

\section{Extended Stream Length (4{,}000 Questions)}
\label{app:extended_stream}

To test whether SECL's calibration improvements are stable over longer streams, we double the stream length from 2{,}000 to 4{,}000 questions (1{,}000 per domain) on Llama~3.2-3B, Gemma~2-2B, and Phi~3.5-Mini (3.8B). Table~\ref{tab:extended_stream} reports the results.

\begin{table}[htbp]
\centering
\small
\resizebox{\columnwidth}{!}{%
\begin{tabular}{llcccc}
\toprule
\textbf{Model} & \textbf{Method} & \textbf{ECE}$\downarrow$ & \textbf{Brier}$\downarrow$ & \textbf{AUROC}$\uparrow$ & \textbf{Acc} \\
\midrule
\multirow{2}{*}{Llama 3.2-3B}
  & Verbalized (4{,}000q) & .190 & .313 & .508 & .577 \\
  & SECL (4{,}000q) & \textbf{.051} & \textbf{.239} & \textbf{.607} & \textbf{.581} \\
\midrule
\multirow{2}{*}{Gemma 2-2B}
  & Verbalized (4{,}000q) & .276 & .328 & \textbf{.548} & .508 \\
  & SECL (4{,}000q) & \textbf{.086} & \textbf{.263} & .500 & \textbf{.512} \\
\midrule
\multirow{2}{*}{Phi 3.5-Mini (3.8B)}
  & Verbalized (4{,}000q) & \textbf{.192} & \textbf{.245} & .593 & .687 \\
  & SECL (4{,}000q) & .259 & .260 & \textbf{.712} & \textbf{.696} \\
\bottomrule
\end{tabular}}%
\caption{4{,}000-question comparison against same-stream verbalized baselines. SECL substantially improves calibration on Llama and Gemma. For Phi, stronger long-stream calibration requires sufficient distillation; with the 4{,}000q run, calibration remains worse than verbalized while discrimination and accuracy improve.}
\label{tab:extended_stream}
\end{table}

\begin{table}[t]
\centering
\small
\resizebox{\columnwidth}{!}{%
\begin{tabular}{llccccc}
\toprule
\textbf{Model} & \textbf{Method} & \textbf{FWD-eq}
  & \textbf{ECE}$\downarrow$ & \textbf{Brier}$\downarrow$
  & \textbf{AUROC}$\uparrow$ & \textbf{Acc} \\
\midrule
\multirow{3}{*}{Llama 3.2-3B}
  & Verbalized & 1          & .170 & .292 & .510 & .576 \\
  & DINCO           & $\sim$10   & .101 & \textbf{.207} & \textbf{.762} & .466 \\
  & SECL (ours)     & 4.6        & \textbf{.050} & .241 & .587 & \textbf{.577} \\
\midrule
\multirow{3}{*}{Phi 3.5-Mini (3.8B)}
  & Verbalized & 1          & .251 & .275 & .600 & \textbf{.667} \\
  & DINCO           & $\sim$10   & \textbf{.110} & \textbf{.212} & \textbf{.749} & .560 \\
  & SECL (ours)     & 2.1        & \textbf{.110} & .251 & .521 & .665 \\
\midrule
\multirow{3}{*}{Gemma 2-2B}
  & Verbalized & 1          & .256 & .314 & .558 & \textbf{.516} \\
  & DINCO           & $\sim$10   & .408 & .410 & .566 & .327 \\
  & SECL (ours)     & 1.8        & \textbf{.056} & \textbf{.254} & .548 & .515 \\
\midrule
\multirow{3}{*}{Llama 3.1-8B}
  & Verbalized & 1          & .225 & .258 & .684 & .644 \\
  & DINCO           & $\sim$10   & .117 & \textbf{.210} & \textbf{.756} & .522 \\
  & SECL (ours)     & 2.8        & \textbf{.083} & .222 & .643 & \textbf{.646} \\
\bottomrule
\end{tabular}}%
\vspace{0.3em}
\par\noindent{\small \emph{FWD-eq}: amortized forward-pass equivalents per question over the full stream (see Section~\ref{app:cost}).}
\caption{SECL vs.\ DINCO across all models.
SECL achieves better calibration (ECE) at lower amortized cost,
while DINCO provides superior discrimination (AUROC) at higher cost.}
\label{tab:dinco}
\end{table}

At 4{,}000 questions, SECL improves over verbalized confidence on Llama (ECE: 0.190$\to$0.051; Brier: 0.313$\to$0.239; AUROC: 0.508$\to$0.607) and Gemma (ECE: 0.276$\to$0.086; Brier: 0.328$\to$0.263), though Gemma shows a modest AUROC trade-off (0.548$\to$0.500). For Phi, SECL improves AUROC/accuracy (0.593$\to$0.712; 0.687$\to$0.696) but remains worse on calibration (ECE/Brier: 0.259/0.260 vs.\ 0.192/0.245), indicating under-distillation.

\section{Additional Open-Ended Evaluation (TruthfulQA-Gen)} \label{app:tqgen}

To reduce the gap between multiple-choice evaluation and open-ended generation, we replace the final domain with the TruthfulQA generation split and keep the rest of the stream unchanged. We report the forward and reversed streams (500 questions per domain, 2{,}000 total) and an extended 4{,}000-question forward stream (1{,}000 per domain) to test long-horizon stability on open-ended answers.

\begin{table}[htbp]
\centering
\small
\resizebox{\columnwidth}{!}{%
\begin{tabular}{llccccc}
\toprule
\textbf{Method} & \textbf{Order} & \textbf{Stream} & \textbf{ECE}$\downarrow$ & \textbf{Brier}$\downarrow$ & \textbf{AUROC}$\uparrow$ & \textbf{Acc} \\
\midrule
Verbalized & Forward & 2k & .195 & .287 & .586 & \textbf{.506} \\
P(True) Norm & Forward & 2k & .173 & .273 & \textbf{.625} & \textbf{.506} \\
SECL (Ours) & Forward & 2k & \textbf{.047} & \textbf{.239} & .618 & .500 \\
SECL (Ours) & Reversed & 2k & \textbf{.047} & .241 & \textbf{.625} & .497 \\
\midrule
Verbalized & Forward & 4k & .179 & .278 & .598 & \textbf{.525} \\
SECL (Ours) & Forward & 4k & \textbf{.061} & \textbf{.237} & \textbf{.625} & .522 \\
\bottomrule
\end{tabular}}%
\caption{TruthfulQA generation-domain evaluation (Llama~3.2-3B).
SECL reduces ECE by 76\% on the 2k forward stream (0.195\,$\to$\,0.047), remains stable under reversed ordering, and at 4k still improves over the same-stream verbalized baseline by 66\% (0.179\,$\to$\,0.061).}
\label{tab:tqgen}
\end{table}

\section{Comparison with DINCO}
\label{app:dinco}

DINCO \citep{wang_calibrating_2025} is a recent inference-time calibration
method that normalizes verbalized confidence against self-generated
distractors and combines it with Self-Consistency.
Unlike SECL, DINCO does not adapt the model; it computes a calibrated
confidence score at each inference step using $\sim$10 forward passes
(beam search + P(True) per candidate + NLI + SC sampling).

SECL and DINCO address different trade-offs.
Across all four models, SECL achieves lower or equal ECE
(Llama~3B: 0.050 vs.\ 0.101; Llama~8B: 0.083 vs.\ 0.117;
Phi: 0.110 vs.\ 0.110; Gemma: 0.056 vs.\ 0.408)
with lower total computational cost and no external models.
DINCO achieves stronger AUROC on all four models
using multiple inference passes and an NLI model,
at the cost of $\sim$10$\times$ inference compute.
The calibration--cost trade-off is visualized in Figure~\ref{fig:pareto_main} (main paper).

Under DINCO's default configuration, accuracy is consistently lower than SECL's
(Llama~3B: 46.6\% vs.\ 57.7\%; Llama~8B: 52.2\% vs.\ 64.6\%;
Phi: 56.0\% vs.\ 66.5\%; Gemma: 32.7\% vs.\ 51.5\%),
as beam-search answer selection yields lower task
accuracy than greedy decoding; using greedy answers with beam-search distractors could mitigate this.
On Gemma, DINCO essentially fails: ECE of 0.408 is worse than
the verbalized baseline (0.256), and accuracy drops to 32.7\%.
Gemma's internal representations appear ill-suited
to DINCO's NLI-based reweighting, whereas SECL's distillation
of the discriminative signal still works (ECE 0.056).

\end{document}